\documentclass[10pt,twocolumn,letterpaper]{article}

\usepackage[pagenumbers]{cvpr} 

\usepackage{algorithm}
\usepackage{algpseudocode}
\usepackage{bm}
\usepackage{booktabs}
\usepackage{mathtools}
\usepackage{url}

\let\oldciteauthor=\citeauthor
\def\citeauthor#1{\hypersetup{citecolor=black}\oldciteauthor{#1}}
\let\oldcite=\cite
\def\cite#1{\hypersetup{citecolor=cvprblue}\oldcite{#1}}

\definecolor{cvprblue}{rgb}{0.21,0.49,0.74}
\usepackage[pagebackref,breaklinks,colorlinks,citecolor=cvprblue]{hyperref}
\title{StableAnimator: High-Quality Identity-Preserving Human Image Animation}

\author{Shuyuan Tu$^{1,2}$ \ \ 
Zhen Xing$^{1,2}$ \ \ 
Xintong Han$^{4}$ \ \ 
Zhi-Qi Cheng$^5$ \ \ 
Qi Dai$^3$ \ \ 
Chong Luo$^{3}$ \ \ 
Zuxuan Wu$^{1,2}$ \vspace{0.05in}\\
{$^1$Shanghai Key Lab of Intell. Info. Processing, School of CS, Fudan University} \\
{$^2$Shanghai Collaborative Innovation Center of Intelligent Visual Computing} \\
{$^3$Microsoft Research Asia}  \quad \quad 
{$^4$Huya Inc.} \quad  \quad 
{$^5$Carnegie Mellon University} \\
{\url{https://francis-rings.github.io/StableAnimator}}
}

\begin{document}

\twocolumn[{
\maketitle
\vspace{-1.2em}
\renewcommand\twocolumn[1][]{#1}
\begin{center}
    \centering
    \includegraphics[width=1\textwidth]{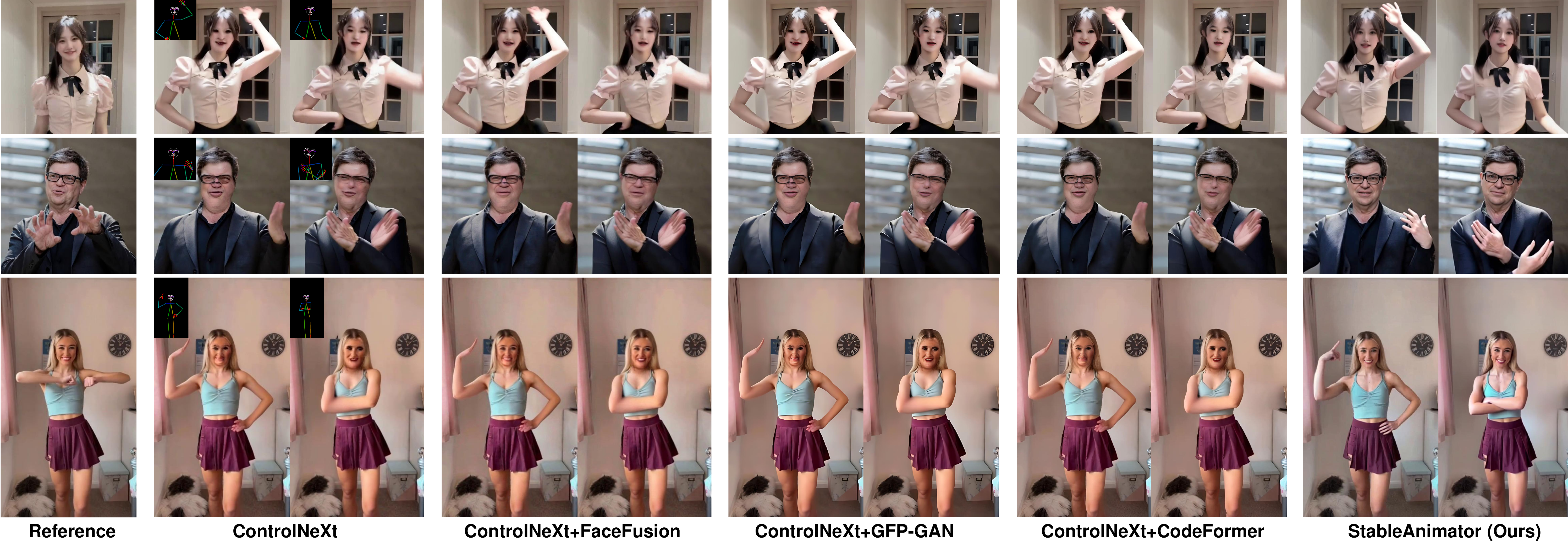}
    \vspace{-0.7cm}
    \captionof{figure}{Pose-driven Human image animations generated by StableAnimator, showing its power to synthesize high-fidelity and ID-preserving videos. FaceFusion~\protect\cite{facefusion} is a face-swapping tool. GFP-GAN~\protect\cite{wang2021gfpgan} and CodeFormer~\protect\cite{zhou2022codeformer} are face restoration models. ControlNeXt~\protect\cite{peng2024controlnext} is the latest open-source animation model.}
    \label{fig:cover}
\end{center}
}
]

\begin{abstract}

Current diffusion models for human image animation struggle to ensure identity (ID) consistency. This paper presents StableAnimator, the first end-to-end ID-preserving video diffusion framework, which synthesizes high-quality videos without any post-processing, conditioned on a reference image and a sequence of poses. Building upon a video diffusion model, StableAnimator contains carefully designed modules for both training and inference striving for identity consistency. In particular, StableAnimator begins by computing image and face embeddings with off-the-shelf extractors, respectively and face embeddings are further refined by interacting with image embeddings using a global content-aware Face Encoder. Then, StableAnimator introduces a novel distribution-aware ID Adapter that prevents interference caused by temporal layers while preserving ID via alignment. During inference, we propose a novel Hamilton-Jacobi-Bellman (HJB) equation-based optimization to further enhance the face quality. We demonstrate that solving the HJB equation can be integrated into the diffusion denoising process, and the resulting solution constrains the denoising path and thus benefits ID preservation. Experiments on multiple benchmarks show the effectiveness of StableAnimator both qualitatively and quantitatively. 

\end{abstract}    
\vspace{-0.5cm}
\section{Introduction}
\label{sec:intro}

Diffusion models~\cite{xing2024aid,dhariwal2021diffusion,ho2020denoising,ho2022cascaded,song2020score,song2020denoising,rombach2022high,meng2021sdedit,hertz2022prompt,tumanyan2023plug,xing2024simda, xing2024survey,weng2024genrec} have achieved remarkable success in image/video generation, significantly inspiring researches in image animation~\cite{wang2024disco, xu2024magicanimate, hu2024animate, zhu2024champ, zhang2024mimicmotion, peng2024controlnext}. In particular, human image animation explores generative models~\cite{Siarohin_2019_NeurIPS, siarohin2021motion,wang2024disco, xu2024magicanimate, hu2024animate, zhu2024champ, zhang2024mimicmotion, peng2024controlnext, wang2024unianimate} 
to animate a reference image conditioned on a sequence of poses through
synthesizing controllable human animation videos, offering diverse applications in entertainment content creation and virtual reality experiences. However, when dealing with pose sequences that exhibit significant motion variation, current approaches suffer from significant distortions and inconsistencies, particularly in facial regions destroying identity information.

To address this issue, there are a number of approaches exploring identity (ID) preservation~\cite{ye2023ip-adapter, wang2024instantid, huang2024consistentid, guo2024pulid} for image generation, yet limited effort has been made for videos. While one could add temporal modeling layers to image diffusion models, doing so would inevitably affect the original spatial priors.
As Image-domain ID-preserving methods rely on stable spatial priors, the interference caused by temporal layers leads to unsatisfactory results.
Thus, for image animation, how to preserve identity information while ensuring video fidelity is extremely challenging. Furthermore, recent animation models~\cite{zhang2024mimicmotion, peng2024controlnext} rely on FaceFusion~\cite{facefusion} for post-processing, which also degrades the quality of animated videos, particularly for facial areas. 

In light of this, we propose StableAnimator, consisting of dedicated modules for both the training and inference to maintain ID consistency for high-quality human image animation. 
StableAnimator first uses off-the-shelf extractors~\cite{deng2019arcface, radford2021learning} to obtain face and image embeddings for the reference image, respectively. 
Face embeddings are further refined by a global content-aware Face Encoder to enable interaction with the reference, enhancing face embeddings' perception of the reference's overall layout, such as backgrounds.
The refined face embeddings are fed to a video diffusion model with a novel distribution-aware ID Adapter that ensures video fidelity while preserving ID clues. 
In particular, diffusion latents perform separate cross-attention with refined face and image embeddings respectively, with their means and variances computed. 
We then use respective means and variances to conduct the alignment between the resulting outputs.
This alignment effectively mitigates interference from the temporal layers by progressively bringing two distributions closer at each step, ensuring ID consistency without compromising video fidelity.

During inference, to further enhance face quality and reduce reliance on post-processing tools, StableAnimator solves
the Hamilton-Jacobi-Bellman (HJB) equation~\cite{bardi1997optimal, peng1992stochastic} for face optimization. 
We find that solving the HJB equation corresponds with the core principles of diffusion denoising. Therefore, we incorporate the HJB equation into the inference process, which allows a controllable variable to guide and constrain the direction of the denoising process. In particular, the solution of HJB is used to update the latents for each denoising step, constraining the denoising path and directing the model toward optimal ID consistency. Since this procedure always adapts to the current distribution of denoised latents, the simultaneous denoising and face optimization effectively eliminates detail distortions. Thus, it can replace the previous over-reliance on third-party post-processing tools, such as face-swapping tools.

As shown in Fig. \ref{fig:cover}, while the latest animation model ControlNeXt~\cite{peng2024controlnext} suffers from dramatic face and body distortion even with face swapping/restoration tools, StableAnimator can accurately animate the reference based on given poses while preserving ID consistency. Experiments on TikTok dataset~\cite{jafarian2021learning} also show that StableAnimator outperforms ControlNeXt by 47.1\% in CSIM~\cite{guo2024liveportrait} while achieving the best result ($140.62$) in FVD.
CSIM is the face similarity between the animated frame and the reference. 

In conclusion, our contributions are as follows:
(1) We propose a global content-aware Face Encoder and a novel distribution-aware ID Adapter to enable the video diffusion model to incorporate face embeddings without compromising video fidelity.
(2) We propose a novel HJB equation-based face optimization method that further enhances face quality while conducting conventional denoising. It is only active in the inference without training any diffusion components. To our knowledge, we are the first to explore video diffusion for end-to-end ID-preserving human image animation.
(3) Experimental results on benchmark datasets show the superiority of our model over the SOTA.
\section{Related Work}
\label{sec:related_work}

\begin{figure*}[t!]
\begin{center}
\includegraphics[width=1\linewidth]{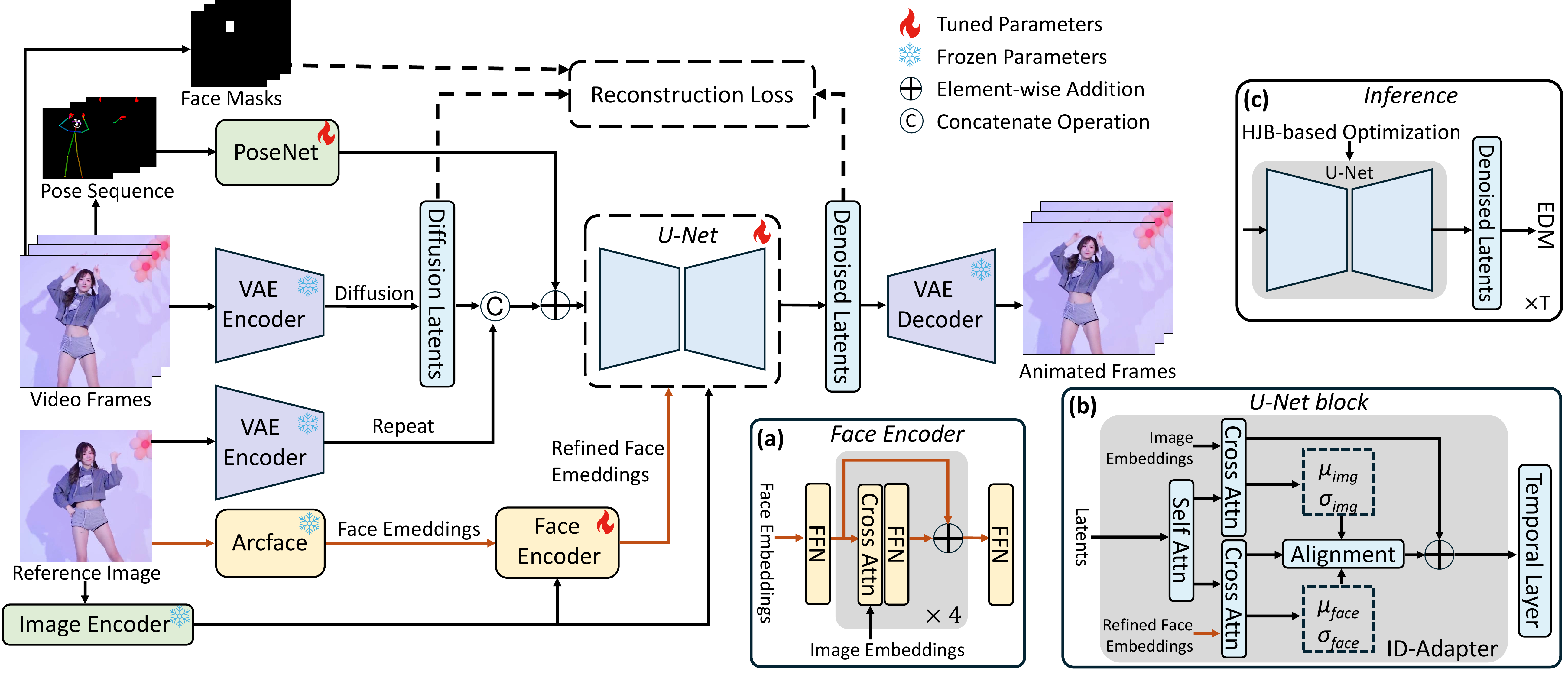}
\end{center}
\vspace{-0.6cm}
   \caption{Architecture of StableAnimator. (a) and (b) refer to the structure of the Face Encoder and each block in the U-Net. Embeddings from the Image Encoder and Face Encoder are injected to each block of U-Net. Given the reference, we extract the image embeddings and face embeddings utilizing Image Encoder and Arcface. 
   The face embeddings are fed into the FaceEncoder to enhance ID. Then, image embeddings and refined face embeddings are injected into the U-Net through the ID Adapter to ensure ID consistency.
   }
\label{fig:framework}
\vspace{-0.5cm}
\end{figure*}

\noindent\textbf{Diffusion for Video Generation.}
Renowned for the capacity in diversity and high-fidelity, diffusion models~\cite{dhariwal2021diffusion,ho2020denoising,ho2022cascaded,nichol2021improved,song2020score,song2020denoising,rombach2022high,meng2021sdedit,hertz2022prompt,tumanyan2023plug,xing2023vidiff} have demonstrated significant success in the video generation. Compared with image generation, video generation requires additional temporal smoothness and temporal consistency. Current video generation models~\cite{singer2022make, guo2023animatediff, wu2023tune, wang2024magicvideo, tu2024motioneditor, videoworldsimulators2024,tu2024motionfollower} tend to add temporal layers to pre-trained image generation diffusion models for joint spatio-temporal modeling. Some researchers replace the diffusion U-Net with the transformer~\cite{peebles2023scalable, yan2021videogpt, yu2023magvit, ma2024latte, bao2024vidu, hong2022cogvideo} for facilitating generative performance. Inspired by previous image animation models~\cite{zhang2024mimicmotion, peng2024controlnext}, we utilize Stable Video Diffusion (SVD~\cite{blattmann2023stable}) as the backbone.

\vspace{0.02in}
\noindent\textbf{ID Consistency Image Generation.}
Studies have explored ID preservation in the image domain. 
LoRA~\cite{hu2021lora} applies a few additional weights for customized dataset training, but it requires individual training for each character, restricting its flexibility.
IP-Adapter-FaceID~\cite{ye2023ip-adapter} attempts to directly separate the cross-attention layers for text features and face features, which potentially introduces the misalignment among features.
PhotoMaker~\cite{li2024photomaker}, FaceStudio~\cite{yan2023facestudio}, and InstantID~\cite{wang2024instantid} present hybrid ID preservation mechanisms for refining face embeddings.
ConsistentID~\cite{huang2024consistentid} designs a facial prompt generator for capturing facial details.
PuLID~\cite{guo2024pulid} introduces contrastive alignment loss and accurate ID loss, ensuring ID fidelity.
However, these models cannot be directly integrated into video diffusion models, as the temporal layers may alter the spatial distribution, resulting in domain mismatching with diffusion latents. This conflict between video fidelity and ID consistency ultimately degrades the quality of animations. By contrast, our StableAnimator can integrate ID information into video diffusion models via a distribution-aware ID Adapter, effectively resolving the above conflict.

\vspace{0.02in}
\noindent\textbf{Pose-guided Human Image Animation.}
Human image animation aims to transfer motion from a given pose sequence to a reference human image. 
Early works~\cite{Siarohin_2019_NeurIPS, siarohin2021motion, huang2021few} basically apply GANs~\cite{goodfellow2020generative} to animate the reference. However, animations of GAN-based models always encounter various artifacts. Recently, some studies have applied diffusion models to this field. 
Disco~\cite{wang2024disco} is the first to use the diffusion model for image animation. 
MagicAnimate~\cite{xu2024magicanimate} and AnimateAnyone~\cite{hu2024animate} both design their reference nets and pose nets to model poses and appearances independently. 
Champ~\cite{zhu2024champ} introduces 3D signal SMPL to enhance controllable capability. Unianimate~\cite{wang2024unianimate} introduces Mamba~\cite{mamba2} to the diffusion model for efficiency. 
MimicMotion~\cite{zhang2024mimicmotion} proposes the regional loss to reduce distortion. 
ControlNeXt~\cite{peng2024controlnext} designs a convolution-based pose net to replace the heavy ControlNet~\cite{zhang2023adding}.
However, previous animation models suffer from face distortion. MimicMotion~\cite{zhang2024mimicmotion} and ControlNeXt~\cite{peng2024controlnext} utilize the third-party face-swapping tool FaceFusion~\cite{facefusion} as post-processing to address this issue, yet this approach can degrade overall video quality. In this paper, our StableAnimator performs end-to-end human image animation that maintains ID consistency without relying on any post-processing tools.
\section{Method}
\label{sec:method}

% \subsection{Architecture Overview}
% \label{sec: architecture_overview}
Illustrated in Fig. \ref{fig:framework}, StableAnimator is based on the commonly used SVD~\cite{blattmann2023stable} following previous works~\cite{zhang2024mimicmotion, peng2024controlnext}. 
A reference image is processed through the diffusion model via three pathways: 
(1) Transformed into a latent code by a frozen VAE Encoder~\cite{kingma2013auto}.
The latent code is duplicated to match video frames, then concatenated with main latents. 
(2) Encoded by the CLIP Image Encoder~\cite{radford2021learning} to obtain image embeddings, which are fed to each cross-attention block of a denoising U-Net and our Face Encoder, respectively, to modulate the synthesized appearance. 
(3) Input to Arcface~\cite{deng2019arcface} to gain face embeddings, which are subsequently refined for further alignment via our Face Encoder.
Refined face embeddings are then fed to the denoising U-Net. 
More details are described in Sec. \ref{sec: face_consistency}. A PoseNet with a similar architecture as AnimateAnyone~\cite{hu2024animate} extracts the features of the pose sequence, which are then added to the noisy latents.

We replace the original input video frames with random noise during inference, while the other inputs stay the same. We propose a novel HJB-equation-based face optimization to enhance ID consistency and eliminate reliance on third-party post-processing tools.
It integrates the solution process of the HJB equation into the denoising, allowing optimal gradient direction toward high ID consistency as detailed in Sec. \ref{sec: optimization}.

\subsection{ID-preserving During Training}
\label{sec: face_consistency}
\textbf{Global Content-aware Face Encoder.} Our goal is to animate the reference image under the guidance of the pose sequence while preserving the ID of the reference image. Directly feeding face embeddings into the U-Net can enrich the diffusion model with face-related information, but lacks awareness of the global context (layout and background) in the reference image before being injected into the U-Net. As a result, ID-irrelevant elements in the reference image bring noise to face modeling, degrading the overall quality of animations.
To address this, we propose a Global Content-Aware Face Encoder, in which the face embeddings go through multiple cross-attention blocks to interact with the reference image embeddings as shown in Fig. \ref{fig:framework}.

\vspace{0.1cm}
\noindent\textbf{Distribution-aware ID Adapter.}
The outputs of the Face Encoder are then fed to our ID Adapter for further alignment to avoid
the distortion of spatial features occurring when directly incorporating image-domain ID-preserving methods~\cite{Siarohin_2019_NeurIPS, guo2024pulid, huang2024consistentid, wang2024instantid} into video diffusion model.
Feature distortion describes the misalignment between face embeddings and spatial diffusion latents, caused by distribution shifts when temporal layers are added at each denoising step.
Image-domain ID-preserving methods rely heavily on a stable spatial distribution of diffusion latents, but temporal layers often alter this distribution, leading to instability in ID preservation.
Such distortion causes a conflict between maintaining high video fidelity and preserving ID consistency. Thus, animated videos often suffer from noticeable blurring effects and can even lose background details. 
The Distribution-aware ID Adapter modifies each spatial layer of the U-Net, as shown in Fig. \ref{fig:framework} (b). Before each temporal modeling, our ID Adapter aligns refined face embeddings with diffusion latents based on their feature distributions, effectively avoiding feature distortion. 

Concretely, following the standard operation of spatial layers in the diffusion model, we first apply spatial self-attention on latents $\bm{z}_{i}$. The latents of the U-Net perform cross-attention with image embeddings $\bm{emb}_{img}$ and refined face embeddings $\bm{emb}_{face}$, respectively:
\begin{equation}\small
\label{eq:cross_attention}
\begin{aligned}
     \bm{z}_{i}&=\mathtt{SAttn}(\bm{z}_{i}), \\
     \bm{z}^{img}_{i}&=\mathtt{CAttn}(\bm{z}_{i}, \bm{emb}_{img}), \\
     \bm{z}^{face}_{i}&=\mathtt{CAttn}(\bm{z}_{i}, \bm{emb}_{face}),
\end{aligned}
\end{equation}
where $\mathtt{SAttn}(\cdot)$ and $\mathtt{CAttn}(\cdot)$ refer to self-attention and cross-attention operations.  
To align $\bm{z}^{img}_{i}$ and $\bm{z}^{face}_{i}$, we enforce $\frac{\bm{z}^{img}_{i}-\bm{\mu}_{img}}{\bm{\sigma}_{img}}=\frac{\bm{z}^{face}_{i}-\bm{\mu}_{face}}{\bm{\sigma}_{face}}$, where $\bm{\mu}_{img/face}$ and $\bm{\sigma}_{img/face}$ refer to the mean and standard deviation of $\bm{z}^{img/face}_{i}$, respectively. If the equation above holds, the feature distributions on both sides are basically in the same domain. Thus, the aligned $\bm{z}^{face}_{i}$ is element-wise added to $\bm{z}^{img}_{i}$ for maintaining ID consistency:
\begin{equation}\small
\label{eq:alignment}
\begin{aligned}
     \bar{\bm{z}}^{face}_{i}&=\frac{\bm{z}^{face}_{i}-\bm{\mu}_{face}}{\bm{\sigma}_{face}}\times\bm{\sigma}_{img}+\bm{\mu}_{img}, \\
     \bar{\bm{z}_{i}}&=\bar{\bm{z}}^{face}_{i}+\bm{z}^{img}_{i}.
\end{aligned}
\end{equation}
The outputs of our ID Adapter $\bar{\bm{z}_{i}}$ are further fed to temporal layers for temporal modeling. 
When spatial distribution is altered by temporal layers, the aligned $\bar{\bm{z}}^{face}_{i}$ remains in the same domain as $\bm{z}^{img}_{i}$, enabling the original $\bm{z}^{face}_{i}$ to reduce reliance on the unstable spatial distribution. Thus, subsequent temporal modeling does not impede the injection of ID information into the U-Net.

\begin{algorithm}[t!]
\caption{Face Optimization ($\sigma(t) = t$ and $s(t) = 1$)}
\label{alg:face_optimization}
\begin{algorithmic}
\small 
\State \textbf{Input:} {$\mathtt{D}_{\theta}(\bm{x}; \bm{\sigma}),t_{i\in \{0, \ldots, N\}}, \bm{\gamma}_{i\in \{0, \ldots, N-1\}}, \bm{y}$} 
    \State \textbf{Sample} $\bm{x}_0 \sim \mathcal{N}(0, t_0^2\bm{I})$ \hfill $\triangleright$ $\mathtt{D}_{\theta}(\bm{x}; \bm{\sigma})$ is a diffusion model
    \State \textbf{For} $i \in \{0, \ldots, N-1\}$ \textbf{do} \hfill $\triangleright$ $t_{i\in \{0, \ldots, N\}}$ are timesteps
        \State \hspace{1em} $\bm{\gamma}_i = 0$ \hfill $\triangleright$ $\bm{\gamma}_{i\in \{0, \ldots, N-1\}}$ are pre-defined factors.
        \State \hspace{1em} \textbf{if} $t_i \in [\bm{S}_{t_{\text{min}}}, \bm{S}_{t_{\text{max}}}]:$ \hfill $\triangleright$ $\bm{y}$ is the reference image.
        \State \hspace{1em} \hspace{1em} $\bm{\gamma}_i = \min \left( \frac{\bm{S}_{\text{churn}}}{N}, \sqrt{2}-1 \right)$
        \State \hspace{1em} \textbf{Sample} $\bm{\epsilon}_i \sim \mathcal{N}(0, \bm{S}_{\text{noise}}^2\bm{I})$
        \State \hspace{1em} $\hat{t}_i = t_i + \bm{\gamma}_i t_i$
        \State \hspace{1em} $\hat{\bm{x}}_i = \bm{x}_i + \sqrt{\hat{t}_i^2 - t_i^2} \bm{\epsilon}_i$
        \State \hspace{1em} $\bm{x}_{\text{pred}}=\mathtt{D}_{\theta}(\hat{\bm{x}}_i; \hat{t}_i)$
        \State \hspace{1em} $\bm{x}_{\text{op}}=\bm{x}_{\text{pred}}.\mathtt{clone}().\mathtt{detach}()$ \hfill $\triangleright$ Starting optimization 
        \State \hspace{1em} $\bm{op}=\mathtt{Adam}([\bm{x}_{\text{op}}], \bm{\eta})$ \hfill $\triangleright$ $\mathtt{Adam}$ optimizer
        \State \hspace{1em} $\bm{x}_{\text{op}}.\text{requires\_grad}=\text{True}$ \hfill $\triangleright$ $\bm{x}_{\text{op}}$ is a HJB variable
        \State \hspace{1em} \textbf{For} $k \in \{1,2, \ldots, 10\}$ \textbf{do} \hfill $\triangleright$ $k$ is the optimization step
        \State \hspace{1em} \hspace{1em} $\bm{f}_{\text{pred}}=\mathtt{Decoder}(\bm{x}_{\text{op}})$ \hfill $\triangleright$ $\mathtt{Decoder}$ is a VAE decoder
        \State \hspace{1em} \hspace{1em} $\bm{loss}=(1-\mathtt{Cos}(\mathtt{Arc}(\bm{f}_{\text{pred}}), \mathtt{Arc}(\bm{y}))).\text{abs}().\text{mean}()$
        \State \hspace{1em} \hspace{1em}  $\bm{op}.\text{zero\_grad}()$
        \State \hspace{1em} \hspace{1em}  $\bm{loss}.\text{backward}(\text{retain\_graph=True})$
        \State \hspace{1em} \hspace{1em}  $\bm{op}.\text{step}()$
        \State \hspace{1em} $\bm{x}_{\text{pred}}=\bm{x}_{\text{op}}$ \hfill $\triangleright$ End of Optimization 
        \State \hspace{1em} $\bm{d}_i = (\hat{\bm{x}}_i-\bm{x}_{\text{pred}})/\hat{t}_i$
        \State \hspace{1em} $\bm{x}_{i+1} = \hat{\bm{x}}_i + (t_{i+1} - \hat{t}_i)\bm{d}_i$
        \State \hspace{1em} \textbf{if} $t_{i+1} \neq 0$:
        \State \hspace{1em} \hspace{1em} $\bm{d}'_i = (\bm{x}_{i+1}-\mathtt{D}_{\theta}(\bm{x}_{i+1}; t_{i+1}))/t_{i+1}$
        \State \hspace{1em} \hspace{1em} $\bm{x}_{i+1} = \hat{\bm{x}}_i + (t_{i+1} - \hat{t}_i) \left( \frac{1}{2} \bm{d}_i + \frac{1}{2} \bm{d}'_i \right)$
    \State \textbf{return} $\bm{x}_N$
\end{algorithmic}
\end{algorithm}

\subsection{ID-preserving During Inference}
\label{sec: optimization}
To improve ID consistency, the latest animation works~\cite{zhang2024mimicmotion, peng2024controlnext} use a third-party face-swapping tool FaceFusion~\cite{facefusion} for post-processing faces. However, animations suffer from overall quality degradation due to excessive reliance on post-processing tools. The reason is that post-processing tools can disrupt the original pixel distribution, as faces generated by third-party tools are clearly not aligned with the domain of original animations. 
To address this issue, inspired by the HJB equation~\cite{bardi1997optimal, peng1992stochastic, chen2023generative}, we propose the HJB Equation-based Face Optimization.
The HJB equation guides optimal variable selection at each moment in a dynamic system to maximize the cumulative reward.
In our setting, this reward refers to ID consistency, which we aim to enhance by integrating the HJB equation with the diffusion denoising process.
The variable refers to the predicted sample by the diffusion model at each denoising iteration.
We first introduce the process of our face optimization and then demonstrate its rationale.

In particular, we optimize the predicted sample $\bm{x}_{\text{pred}}$ by minimizing the face similarity distance between $\bm{x}_{\text{pred}}$ and the reference before employing denoising (EDM~\cite{karras2022elucidating}) at each step. 
The details are in the Algorithm \ref{alg:face_optimization}, following the structure of the Algorithm 2 in the EDM paper~\cite{karras2022elucidating}. 
$\bm{S}_{\text{noise}}$,$ \bm{S}_{\text{churn}}$, $\bm{S}_{t_{\text{min}}}$, and 
$\bm{S}_{t_{\text{max}}}$ are the pre-defined values of EDM. $\mathtt{Arc}(\cdot)$ and $\bm{\eta}$ are Arcface~\cite{deng2019arcface} and a learning rate. We employ our optimization to refine the prediction of the diffusion regarding the face similarity with the reference.

The optimized $\bm{x}_{\text{pred}}$ can steer the denoising process forward in a way that maximizes ID consistency. As our optimization relies on the current distribution of denoised latents from diffusion, this parallel operation of denoising and optimizing ID consistency effectively reduces detail distortions, enhancing face quality.

Furthermore, we prove that the solving process of the HJB equation~\cite{bardi1997optimal, peng1992stochastic, chen2023generative} can be integrated with the diffusion denoising process, as demonstrated below.
The basic HJB Equation can be described as:
\begin{equation}\small
\label{eq:basic_HJB}
\begin{aligned}
     \frac{\partial \mathtt{V}(\bm{x}, t)}{\partial t}+\mathtt{max}_{c}[\mathtt{f}(\bm{x}, \bm{c})+\frac{\partial \mathtt{V}(\bm{x}, t)}{\partial \bm{x}}\cdot\mathtt{g}(\bm{x}, \bm{c})]=0,
\end{aligned}
\end{equation}
where $\mathtt{V}(\bm{x}, t)$ refers to the value function, representing the minimum cost from state $\bm{x}$ at time $t$. $\mathtt{f}(\bm{x}, \bm{c})$ is the immediate cost under the condition $\bm{c}$ in state $\bm{x}$. $\mathtt{g}(\cdot)$ depicts the system dynamics. 
In our settings, the condition $\bm{c}$ indicates the face-aware variable.
Following the previous work~\cite{chen2023generative}, the solving process is formulated as:
\begin{equation}\small
\label{eq:particular_HJB}
\begin{aligned}
     \mathtt{min}_{\bm{c}_{t}}\int_{0}^{1}\frac{1}{2}\left \|\bm{c}_{t}\right \|_{2}^{2}dt+\frac{\bm{r}}{2}\left \|\bm{X}_{1}-\bm{x}_{1}\right \|_{2}^{2}, \bm{X}_{1}\sim \bm{p}_{data}, 
\end{aligned}
\end{equation}
s.t. $d\bm{X}_{t}=\bm{c}_{t}dt$ and $\bm{X}_{0}=\bm{x}_{0}$ (Gaussian noise). $\bm{r}$ is the terminal cost coefficient. 
In our work, we normalize denoising timesteps ${t}'$ (from $\bm{T}$ to $0$) to $[0, 1]$ and set $t=1-{t}'$.
$\bm{T}$ is the maximum denoising timestep. $\bm{X}_{t}$ and $\bm{x}_{t}$ refer to the groundtruth sample and the predicted sample by the model. 
Thus, $\bm{x}_{\text{pred}}$ in Algorithm \ref{alg:face_optimization} is equivalent to $\bm{x}_{1}$.
Following the Pontryagin Maximum Principle~\cite{kirk2004optimal}, we can construct the Hamiltonian equation:
\begin{equation}\small
\label{eq:hamiltonian}
\begin{aligned}
     \mathtt{H}(t,\bm{X},\bm{c}_{t},\bm{\gamma})=-\frac{1}{2}\left \|\bm{c}_{t}\right \|_{2}^{2}+\bm{\gamma}\bm{c}_{t},
\end{aligned}
\end{equation}
where $\bm{\gamma}$ refers to a coefficient.
To minimize Eq. \ref{eq:hamiltonian}, we set $\frac{\partial \mathtt{H}}{\partial \bm{c}_{t}}=0$. The optimal Hamiltonian is described as:
\begin{equation}\small
\label{eq:optimized_hamiltonian}
\begin{aligned}
     \mathtt{H}^{*}=\mathtt{H}(t,\bm{X},{\bm{c}}^{*}_{t},\bm{\gamma})=\frac{1}{2}\bm{\gamma}^{2}, \text{where } {\bm{c}}^{*}_{t}=\bm{\gamma}. 
\end{aligned}
\end{equation}
Then we solve the Hamiltonian equation of motion:
\begin{equation}\small
\label{eq:motion}
\begin{aligned}
     \frac{d\bm{X}_{t}}{dt}&=\frac{\partial \mathtt{H}^{*}}{\partial \bm{\gamma}}=\bm{\gamma}, \\
     \frac{d\bm{\gamma}}{dt}&=\frac{\partial \mathtt{H}^{*}}{\partial \bm{X}}=0.
\end{aligned}
\end{equation}
At the final step $t=1$, from Eq. \ref{eq:particular_HJB} and Eq. \ref{eq:hamiltonian}, we can obtain $\bm{\gamma}_{1}=-\bm{r}\cdot(\bm{X}_{1}-\bm{x}_{1})$. From Eq. \ref{eq:motion}, we can see that $\bm{\gamma}$ is a variable independent of $t$, thereby obtaining $\bm{\gamma}=\bm{\gamma}_{1}=-\bm{r}\cdot(\bm{X}_{1}-\bm{x}_{1})$. We can also get $\bm{X}_{t}=\bm{X}_{0}+\bm{\gamma}t$ $\rightarrow$ $\bm{X}_{1}=\bm{X}_{0}+\bm{\gamma}$ and $\bm{X}_{0}=\bm{X}_{t}-\bm{\gamma}t$. We then obtain ${\bm{c}}^{*}_{t}$:
\begin{equation}\small
\label{eq:optimal_c}
\begin{aligned}
    &\bm{X}_{1}=\bm{X}_{0}+\bm{\gamma}=\bm{X}_{t}-\bm{\gamma}t+\bm{\gamma} \\ 
     \rightarrow&\quad \bm{\gamma}=-\bm{r}\cdot(\bm{X}_{1}-\bm{x}_{1})=-\bm{r}\cdot(\bm{X}_{t} - \bm{\gamma}t + \bm{\gamma} -\bm{x}_{1}), \\
     \rightarrow&\quad {\bm{c}}^{*}_{t}=\bm{\gamma}=\frac{\bm{r}(\bm{x}_{1}-\bm{X}_{t})}{1+\bm{r}(1-t)}.
\end{aligned}
\end{equation}
When $\bm{r} \to \infty$, following Eq. \ref{eq:particular_HJB} ($d\bm{X}_{t}=\bm{c}_{t}dt$) and certainty equivalence~\cite{fleming2012deterministic,chen2023generative} (the stochastic case), we have
\begin{equation}\small
\label{eq:extreme_formula}
\begin{aligned}
     d\bm{X}_{t}=\frac{\bm{x}_{1}-\bm{X}_{t}}{1-t}dt+d\bm{w}_{t},
\end{aligned}
\end{equation}
where $\bm{w}_{t}$ is Brownian motion~\cite{chen2023generative}.
According to EDM~\cite{karras2022elucidating} in SVD~\cite{blattmann2023stable}, where $\bm{X}_{{t}'}=\bm{X}_{data}+{t}'\bm{\varepsilon}$ and $\bm{X}_{data}\sim \bm{p}_{data}$, the current state $\bm{X}_{{t}'}$ is converted to $\bm{X}_{t}=\bm{X}_{1}+(1-t)\bm{\varepsilon}$ in our settings. We use the following Tweedie’s formula~\cite{efron2011tweedie} 
\begin{equation}\small
\label{eq:tweedie}
\begin{aligned}
     \mathtt{E}[\bm{\theta}|\bm{x}]=\bm{x}+\bm{\sigma}^{2}\cdot\nabla\log\mathtt{p}(\bm{x}),
\end{aligned}
\end{equation}
where $\bm{x}|\bm{\theta}\sim\mathcal{N}(\bm{\theta}, \bm{\sigma}^{2})$ and $\mathtt{p}(\cdot)$ is the marginal density of $\bm{x}$, to reform $\bm{X}_{1}$:
\begin{equation}\small
\label{eq:approximation}
\begin{aligned}
     \bm{X}_{1}=\mathtt{E}[\bm{X}_{1}|\bm{X}_{t}]=\bm{X}_{t}+(1-t)^{2}\nabla\log\mathtt{p}(\bm{X}_{t}).
\end{aligned}
\end{equation}
$\bm{x}_{1}$ aims to approximate $\bm{X}_{1}$. Thus, we substitute Eq. \ref{eq:approximation} in Eq. \ref{eq:extreme_formula} for obtaining the ultimate formula:
\begin{equation}\small
\label{eq:final}
\begin{aligned}
     d\bm{X}_{t}&=\frac{\bm{X}_{t}+(1-t)^{2}\nabla\log\mathtt{p}(\bm{X}_{t})-\bm{X}_{t}}{1-t}dt+d\bm{w}_{t} \\
     &=(1-t)\cdot\nabla\log\mathtt{p}(\bm{X}_{t})dt+d\bm{w}_{t}.
\end{aligned}
\end{equation}

It is evident that Eq. \ref{eq:final} and SDE formulation~\cite{song2020score} are structurally the same, thus we can seamlessly incorporate the solution process of the HJB equation into the diffusion denoising for ID preservation.

\begin{table*}[t!]\small
\caption{Quantitative comparisons on TikTok dataset and Unseen100. Mem refers to GPU memory when manipulating 16 frames ($576\times1024$).  In the table elements $a$ / $b$, $a$, and $b$ refer to the result on the TikTok dataset and Unseen100, respectively. We reference competitors' results on the TikTok dataset from their papers, with $-$ indicating missing reports.
}
\vspace{-0.25in}
\begin{center}
\renewcommand\arraystretch{1.1}
\scalebox{0.9}{
\begin{tabular}{l|cccccc|c|c}
\toprule
Model          & L1 (E-4)$\downarrow$               & PSNR~\cite{hore2010image}$\uparrow$          & PSNR*~\cite{wang2024disco}$\uparrow$           & SSIM$\uparrow$           & LPIPS$\downarrow$          & CSIM~\cite{guo2024liveportrait}$\uparrow$           & FVD$\downarrow$             & Mem$\downarrow$           \\ \midrule
MRAA~\cite{siarohin2021motion}           & 3.21 / 3.62          & - / 26.62              & 18.14 / 17.28          & 0.672 / 0.692          & 0.296 / 0.313          & 0.248 / 0.221          & 284.82 / 540.35          & \textbf{5.4G}          \\ \midrule
DisCo~\cite{wang2024disco}          & 3.78 / 3.74          & 29.03 / 25.23          & 16.55 / 15.21          & 0.668 / 0.702          & 0.292 / 0.302          & 0.315 / 0.267          & 292.80 / 544.64          & 18.7G         \\
MagicAnimate~\cite{xu2024magicanimate}   & 3.13 / 3.23          & 29.16 / 27.03          & - / 17.11              & 0.714 / 0.746          & 0.239 / 0.264          & 0.462 / 0.338           & 179.07 / 398.94          & 20.84G        \\
AnimateAnyone~\cite{hu2024animate}  & - / 3.15                & 29.56 / 27.14          & - / 17.14              & 0.718 / 0.759          & 0.285 / 0.251          & 0.457 / 0.316          & 171.90 / 383.45          & 11.18G        \\
Champ~\cite{zhu2024champ}          & 2.94 / 3.02          & 29.91 / 27.78          & - / 17.35              & 0.802 / 0.772          & 0.231 / 0.234          & 0.350 / 0.304          & 160.82 / 373.77          & 13.2G         \\
Unianimate~\cite{wang2024unianimate}     & \textbf{2.66} / 2.82 & 30.77 / 27.46          & 20.58 / 18.64          & \textbf{0.811} / 0.778 & \textbf{0.231} / 0.253 & 0.479 / 0.347          & 148.06 / 394.32          & 6.11G         \\
MimicMotion~\cite{zhang2024mimicmotion}    & 5.85 / 3.55          & - / 22.94              & 14.44 / 13.97          & 0.601 / 0.733          & 0.416 / 0.370          & 0.262 / 0.242          & 326.57 / 604.13          & 8.6G          \\
ControlNeXt~\cite{peng2024controlnext}   & 6.20 / 2.90          & - / 25.28              & 13.83 / 14.84          & 0.615 / 0.743          & 0.416 / 0.262          & 0.360 / 0.264          & 326.57 / 389.45          & 12.23G        \\ \midrule
\textbf{StableAnimator} & 2.87 / \textbf{2.71}          & \textbf{30.81} / \textbf{28.85} & \textbf{20.66} / \textbf{18.85} & 0.801 / \textbf{0.784}          & 0.232 / \textbf{0.223}          & \textbf{0.831} / \textbf{0.805} & \textbf{140.62} / \textbf{349.94} & 12.50G        \\ \bottomrule
\end{tabular}
}
\end{center}
\vspace{-0.2in}
\label{table:quantitative_comparisons}
\end{table*}

\subsection{Training}
\label{sec: training}
As illustrated in Fig. \ref{fig:framework}, we use the reconstruction loss to train our model, with trainable components including a U-Net, a FaceEncoder, and a PoseNet. 
We introduce face masks $\bm{M}$, extracted by ArcFace~\cite{deng2019arcface} from the input video frames to enhance the modeling of face regions:
\begin{equation}\small
\label{eq:loss}
\begin{aligned}
     \mathcal{L}=\mathbb{E}_{\varepsilon}(\left \| (\bm{z}_{gt}-\bm{z}_{\varepsilon})\odot  (1+\bm{M})  \right \|^{2}),
\end{aligned}
\end{equation}
where $\bm{z}_{gt}$ and $\bm{z}_{\varepsilon}$ refer to diffusion latents and denoised latents in Fig. \ref{fig:framework}, respectively.

\section{Experiments}
\label{sec:experiments}

\subsection{Implementation Details}
Since previous works do not open-source their training datasets, we collect 3K videos (60-90 seconds long) from the internet to train our model. We utilize DWPose~\cite{yang2023effective} and Arcface~\cite{deng2019arcface} to extract skeleton poses and face embeddings/masks.
Following previous works~\cite{hu2024animate, xu2024magicanimate, zhu2024champ, wang2024disco, wang2024unianimate}, we evaluate our model on TikTok dataset~\cite{jafarian2021learning}. 
We conduct additional experiments on 100 unseen videos, referred to the Unseen100 dataset, selected from the internet to assess the robustness of our model.
Following recent animation models~\cite{zhang2024mimicmotion,peng2024controlnext}, the U-Net utilizes pre-trained weights of SVD~\cite{blattmann2023stable}, while the PoseNet and Face Encoder are trained from scratch. 
Our ID-Adapter uses pre-trained weights of spatial cross-attention blocks in SVD. 
Our model is trained for 20 epochs on 4 NVIDIA A100 80G GPUs, with a batch size of 1 per GPU. The learning rate is set to 1$e$-5. 

\subsection{Comparison with State-of-the-Art Methods}

\textbf{Quantitative results.}
We compare with recent human image animation models, including GAN-based models (MRAA~\cite{siarohin2021motion}) and diffusion-based models (DisCo~\cite{wang2024disco}, AnimateAnyone~\cite{hu2024animate}, MagicAnimate~\cite{xu2024magicanimate}, Champ~\cite{zhu2024champ}, Unianimate~\cite{wang2024unianimate}, MimicMotion~\cite{zhang2024mimicmotion}, ControlNeXt~\cite{peng2024controlnext}). Based on previous studies that assess quantitative results using the self-driven and reconstruction approach, we perform quantitative comparisons with the above competitors on the TikTok dataset~\cite{jafarian2021learning} and Unseen100, comprising complex motion and appearance information. 
Notably, all competitors are trained on our dataset before evaluating on Unseen100 to ensure a fair comparison.
The results are shown in Table \ref{table:quantitative_comparisons}. CSIM~\cite{guo2024liveportrait} evaluates the cosine similarity between the facial embeddings of two images.
We observe that our StableAnimator surpasses all competitors regarding face quality (CSIM) and video fidelity (FVD) while maintaining relatively high single-frame quality. Specifically, StableAnimator outperforms the leading competitor, Unianimate, by 36.9\% and 45.8\% in CSIM across two datasets, without sacrificing video fidelity and single-frame quality.

\begin{figure*}[t!]
\begin{center}
\includegraphics[width=0.98\linewidth]{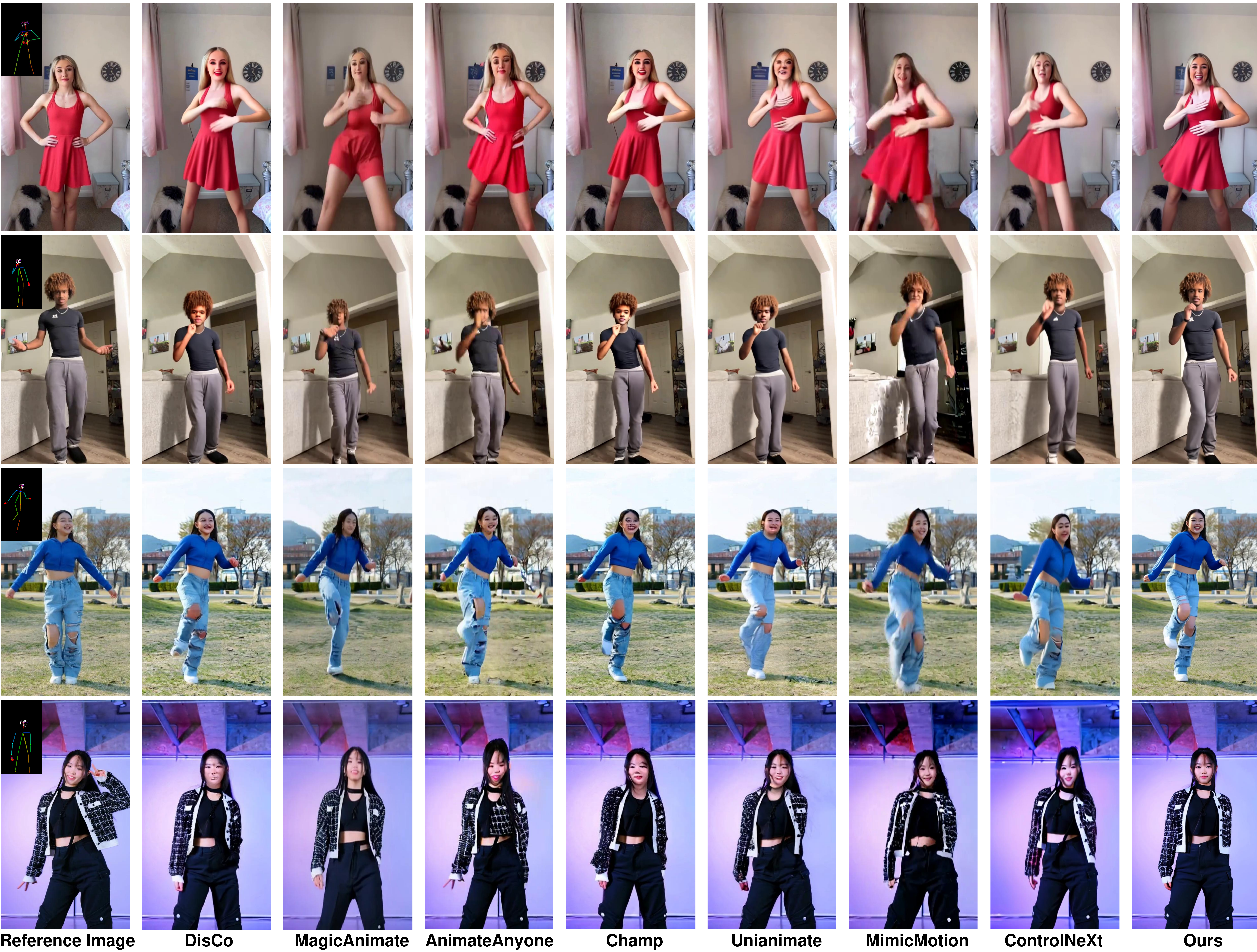}
\end{center}
\vspace{-0.65cm}
   \caption{Qualitative comparisons with state-of-the-art methods. More examples can be found in the supplementary material.}
\label{fig:comparison}
\vspace{-0.5cm}
\end{figure*}

\noindent\textbf{Qualitative Results.}
The qualitative results are shown in Fig. \ref{fig:comparison}. Notably, qualitative results in the paper are in the cross-ID setting~\cite{zhu2024champ}. Disco~\cite{wang2024disco}, MagicAnimate~\cite{xu2024magicanimate}, AnimateAnyone~\cite{hu2024animate}, and Champ~\cite{zhu2024champ} exhibit face/body distortion and clothing changes, while Unianimate~\cite{wang2024unianimate} accurately modifies the reference motion, and MimicMotion~\cite{zhang2024mimicmotion} and ControlNeXt~\cite{peng2024controlnext} effectively preserve clothing details. However, all competitors struggle to maintain reference identities. In contrast, our StableAnimator accurately animates images based on the given pose sequences while preserving reference identities, highlighting the superiority of our model in identity retention and in generating precise, vivid animations.

\subsection{Ablation Study}

\begin{figure}[t!]
\begin{center}
\includegraphics[width=1\linewidth]{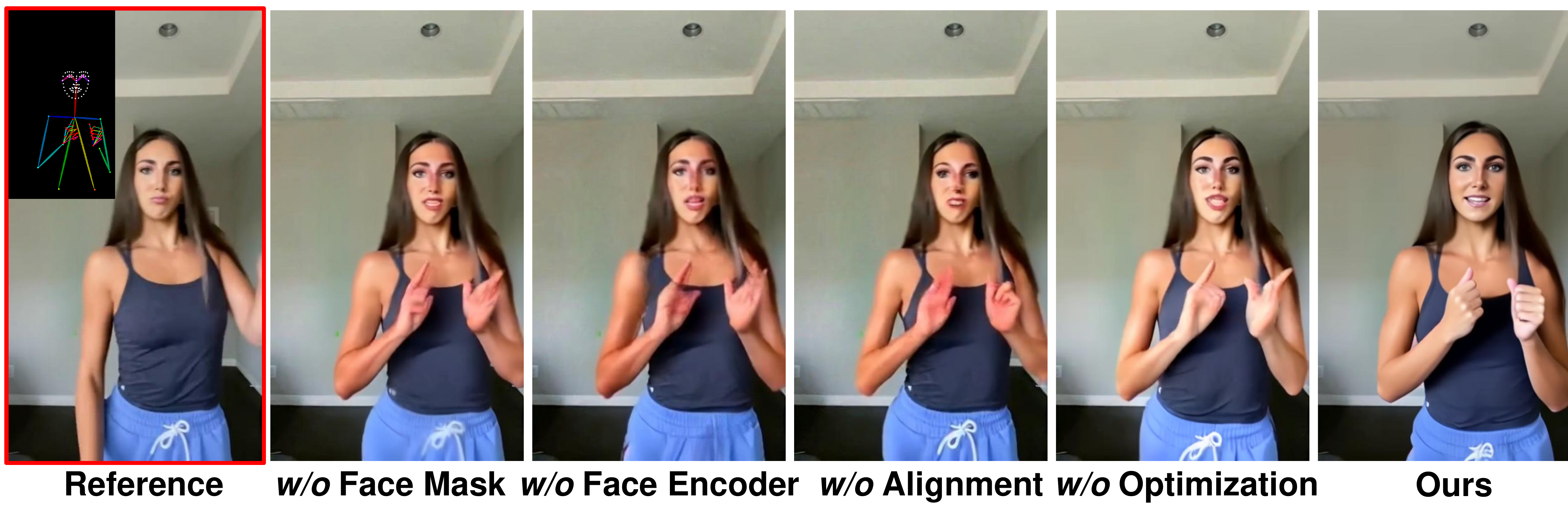}
\end{center}
\vspace{-0.6cm}
   \caption{Ablations on core components of StableAnimator. 
   }
\label{fig:ablation_core}
\vspace{-0.40cm}
\end{figure}

\textbf{ID Consistency Preservation.}
We conduct an ablation study to demonstrate the contributions of core components in StableAnimator, as shown in Table \ref{table:ablation_core} and Fig. \ref{fig:ablation_core}. Notably, all quantitative ablation studies are on the Unseen100 dataset.
We can see that removing the core components significantly degrades performance, particularly in face-related regions (CSIM), highlighting that our components significantly enhance both video fidelity and single-frame quality while preserving high ID consistency.

\begin{figure}[t!]
\begin{center}
\includegraphics[width=1\linewidth]{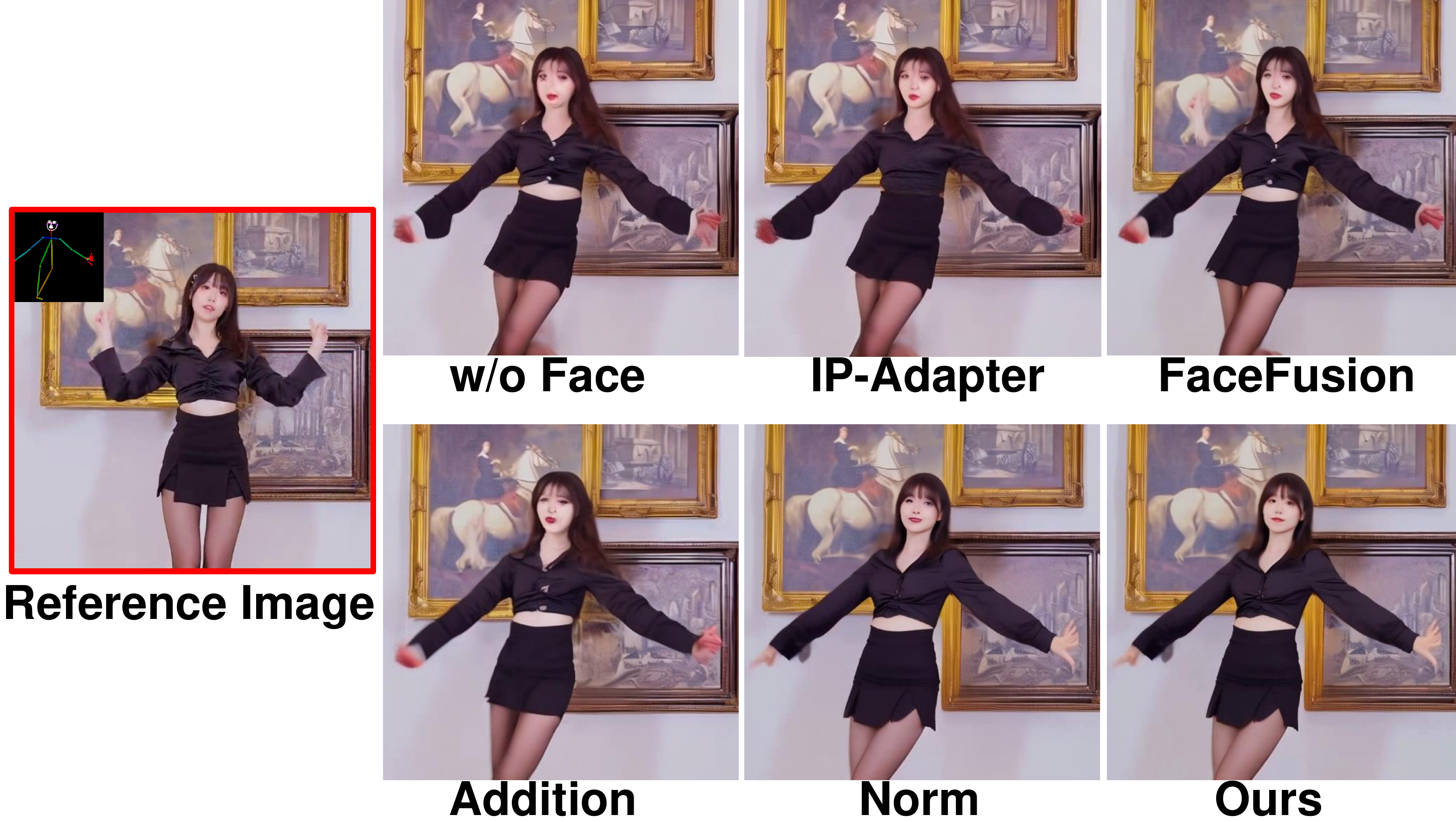}
\end{center}
\vspace{-0.6cm}
   \caption{Ablation study on face enhancement strategies.
   }
\label{fig:ablation_face}
\vspace{-0.35cm}
\end{figure}

We further conduct an ablation study regarding current face enhancement approaches, as shown in Table \ref{table:ablation_face} and Fig. \ref{fig:ablation_face}. We replace our components with the commonly used IP-Adapter and FaceFusion. By analyzing the results, we can gain the following observations: (1) IP-Adapter can improve the ID consistency, while the video fidelity and single-frame quality dramatically degrade. The plausible reason is that directly inserting the IP-Adapter hinders its ability to adapt to spatial representation distribution variations during temporal modeling, thereby deteriorating the capacity of the video diffusion model. (2) The third-party post-processing face-swapping tool FaceFusion refines the face quality but relatively degrades the video fidelity. The underlying reason is that the third-party post-processing operates in a different domain from the diffusion model, leading to a loss of semantic details and disrupting video fidelity. (3) StableAnimator can significantly refine the face quality while maintaining high video fidelity since our model remains in the same domain as the video diffusion model due to the distribution-aware end-to-end pipeline.

\begin{table}[t!]\small
\caption{Ablation study on core components. Face Masks and Alignment refer to face masks in the loss and distribution alignment of our ID Adapter.
}
\vspace{-0.25in}
\begin{center}
\renewcommand\arraystretch{1.1}
\scalebox{0.75}{
\begin{tabular}{lccccc|c}
\toprule
Model            & L1$\downarrow$                          & PSNR$\uparrow$                      & SSIM$\uparrow$                      & LPIPS$\downarrow$                     & CSIM$\uparrow$           & FVD$\downarrow$                        \\ \midrule
\textit{w/o} Face Masks   & 3.01E-4                     & 24.10                     & 0.665                     & 0.281                     & 0.639          & 382.25                     \\
\textit{w/o} Face Encoder & 3.08E-4                     & 22.25                     & 0.674                     & 0.282                     & 0.594          & 385.91                     \\
\textit{w/o} Alignment    & 3.11E-4 & 23.45 & 0.713 & 0.276 & 0.716          & 412.52 \\
\textit{w/o} Optimization & 2.86E-4 & 27.17 & 0.769 & 0.245 & 0.782          & 365.43 \\ \midrule
Ours             & \textbf{2.71E-4}            & \textbf{28.85}            & \textbf{0.784}            & \textbf{0.223}            & \textbf{0.805} & \textbf{349.94}            \\ \bottomrule
\end{tabular}
}
\end{center}
\label{table:ablation_core}
\vspace{-0.3in}
\end{table}

\begin{table}[t!]\small
\caption{Ablation study on face enhancement methods. \textit{w/o} Face refers to the exclusion of any face-related strategies.
}
\vspace{-0.25in}
\begin{center}
\renewcommand\arraystretch{1.1}
\scalebox{0.8}{
\begin{tabular}{lccccc|c}
\toprule
Model      & L1$\downarrow$                          & PSNR$\uparrow$                      & SSIM$\uparrow$                      & LPIPS$\downarrow$                     & CSIM$\uparrow$           & FVD$\downarrow$                        \\ \midrule
\textit{w/o} Face   & 2.83E-4                     & 26.75                     & 0.741                     & 0.264                     & 0.324          & 371.38                     \\
IP-Adapter~\cite{ye2023ip-adapter} & 3.88E-4                     & 18.86                     & 0.672                     & 0.287                     & 0.511          & 484.77                     \\
FaceFusion~\cite{facefusion} & 3.31E-4 & 23.05 & 0.734 & 0.265 & 0.798          & 405.16 \\ \midrule
Ours       & \textbf{2.71E-4}            & \textbf{28.85}            & \textbf{0.784}            & \textbf{0.223}            & \textbf{0.805} & \textbf{349.94}            \\ \bottomrule
\end{tabular}
}
\end{center}
\label{table:ablation_face}
\vspace{-0.3in}
\end{table}

\noindent \textbf{Feature Alignment.} We conduct a comparison between our distribution alignment in the ID-Adapter and other types of feature injection, as shown in Table \ref{table:ablation_alignment} and Fig. \ref{fig:ablation_face}. Norm refers to $\bar{\bm{z}}_{i}^{face}$=$\frac{\bm{z}_{i}^{face}-\bm{\mu}_{face}}{\bm{\sigma}_{face}}$. We can see that Addition and Norm fail to eliminate the interference of spatial feature distortion after temporal modeling, thereby achieving suboptimal results. By contrast, our alignment integrates the mean and standard deviation from both cross-attention features, significantly mitigating the impact of feature distortion.

\begin{table}[t!]\small
\caption{Ablation study on the distribution-based alignment. Addition and Norm refer to element-wise addition and normalization.
}
\vspace{-0.25in}
\begin{center}
\renewcommand\arraystretch{1.1}
\scalebox{0.8}{
\begin{tabular}{lccccc|c}
\toprule
Model                 & L1$\downarrow$               & PSNR$\uparrow$           & SSIM$\uparrow$           & LPIPS$\downarrow$          & CSIM$\uparrow$           & FVD$\downarrow$             \\ \midrule
Addition & 3.11E-4          & 23.45          & 0.713          & 0.276          & 0.716          & 412.52          \\
Norm         & 2.73E-4          & 26.67          & 0.758          & 0.257          & 0.776          & 382.49          \\ \midrule
Ours                  & \textbf{2.71E-4} & \textbf{28.85} & \textbf{0.784} & \textbf{0.223} & \textbf{0.805} & \textbf{349.94} \\ \bottomrule
\end{tabular}
}
\end{center}
\label{table:ablation_alignment}
\vspace{-0.2in}
\end{table}

\begin{figure}[t!]
\begin{center}
\includegraphics[width=1\linewidth]{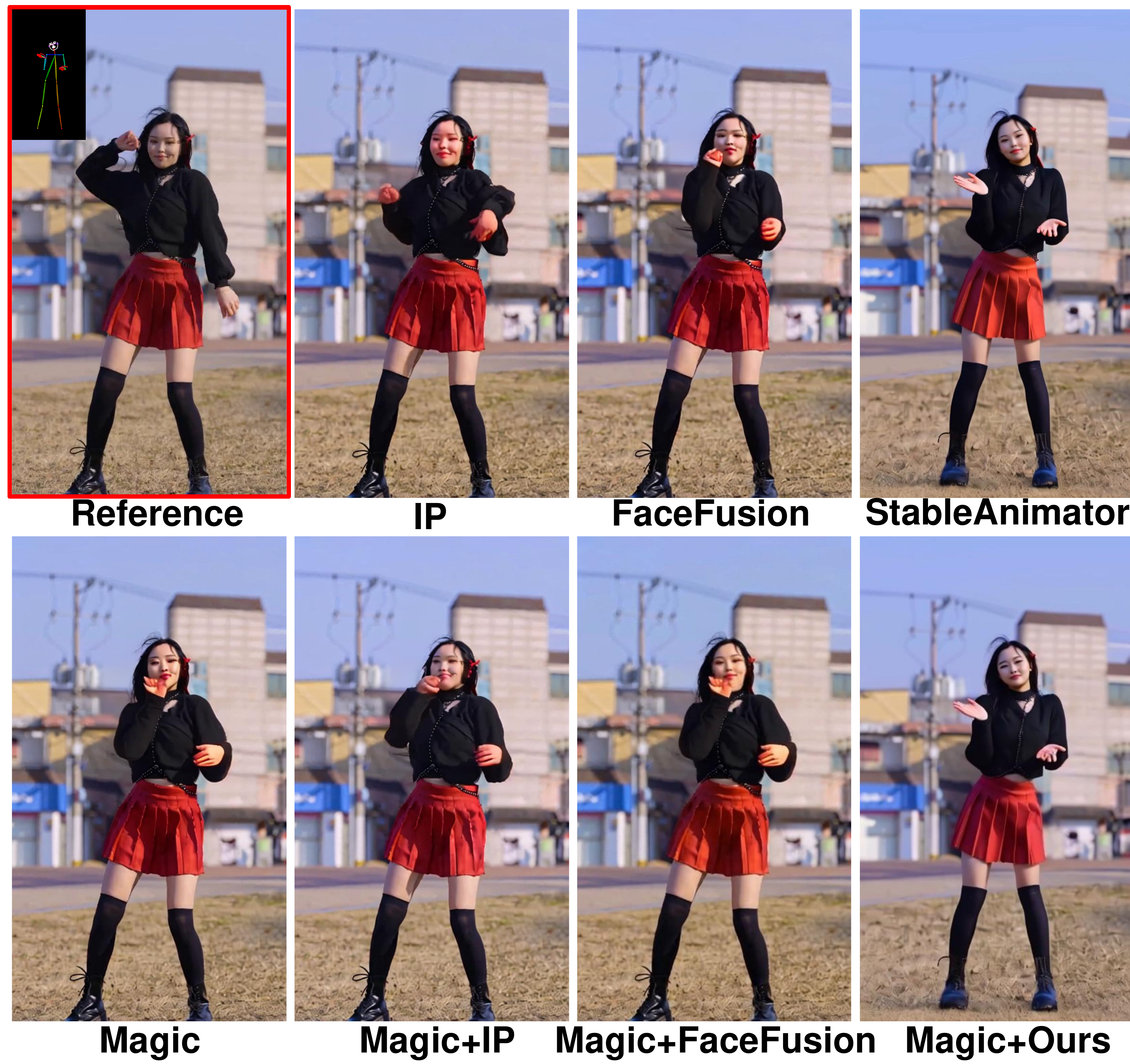}
\end{center}
\vspace{-0.6cm}
   \caption{Ablation study on different backbones.
   }
\label{fig:ablation_opt}
\vspace{-0.5cm}
\end{figure}

\noindent\textbf{Face Optimization.} To validate the significance of our face optimization strategy, we conduct an ablation regarding different diffusion backbones. The results are in Table \ref{table:ablation_opt} and Fig. \ref{fig:ablation_opt}. MagicAnimate is based on SD~\cite{rombach2022high}+AnimateDiff~\cite{guo2023animatediff}. We have the following observations: (1) Common face enhancement strategies (IP-Adapter and FaceFusion) also degrade the video fidelity and single-frame quality of MagicAnimate, indicating that spatial feature distortion indeed occurs across different diffusion-based backbones. 
(2) Magic+Opt boosts overall performance, showing that our face optimization enhances the diffusion model even without any explicit introduction of face-related adapters. The results of Magic+IP+Opt indicate that our optimization can mitigate the deterioration in fidelity due to the introduction of IP-Adapter while improving face quality to some extent.  
(3) The last two rows of Table \ref{table:ablation_opt} show that our face optimization can still work in the different diffusion-based backbone.

\begin{table}[t!]\small
\caption{Ablation study on the optimization. Magic, IP, ID, FE, and Opt refer to MagicAnimate, IP-Adapter, our ID Adapter, our Face Encoder, and our Optimization, respectively.
}
\vspace{-0.25in}
\begin{center}
\renewcommand\arraystretch{1.1}
\scalebox{0.75}{
\begin{tabular}{lccccc|c}
\toprule
Model                                    & L1$\downarrow$               & PSNR$\uparrow$           & SSIM$\uparrow$           & LPIPS$\downarrow$          & CSIM$\uparrow$           & FVD$\downarrow$             \\ \midrule
Magic+IP                          & 3.85E-4          & 23.14          & 0.689          & 0.286          & 0.541          & 836.33          \\
Magic+FaceFusion                  & 3.31E-4          & 26.42          & 0.725          & 0.268          & 0.796          & 412.40          \\
Magic+Opt                & 3.02E-4          & 27.56          & 0.762          & 0.258          & 0.480          & 381.61          \\
Magic+IP+Opt             & 3.61E-4          & 26.12          & 0.714          & 0.279          & 0.624          & 754.34          \\
Magic+FE+ID              & 2.85E-4          & 27.89          & 0.767          & 0.248          & 0.775          & 376.43          \\
Magic+FE+ID+Opt & \textbf{2.69E-4} & \textbf{28.13} & \textbf{0.775} & \textbf{0.241} & \textbf{0.798} & \textbf{355.23} \\ \bottomrule
\end{tabular}
}
\end{center}
\label{table:ablation_opt}
\vspace{-0.2in}
\end{table}

\subsection{Applications and User Study}
\noindent\textbf{Long Animation.} We conduct a qualitative comparison between our StableAnimator and current animation models in long animation generation. Inspired by MimicMotion~\cite{zhang2024mimicmotion}, we follow the same pipeline for synthesizing long videos. The results are shown in Sec. \ref{long_animation} of the Supp. Each pose sequence contains over 300 frames with complex motion, while the references include intricate details of appearances and backgrounds. The results show that competitors suffer from blurry noises and face distortion. By contrast, our model can effectively handle long human image animation in high fidelity while preserving identities.

\noindent\textbf{Multi-Person Animation.} We experiment on multi-person animation, as shown in Sec. \ref{multiple_person} of the Supp. We can see that our model is capable of animating multiple individuals. 

\noindent\textbf{User Study.}
We conduct a user study on 30 selected videos to evaluate the human preference between our StableAnimator and other competitors. The participants are basically university students and faculties. In each case, participants are first presented with the reference image and the pose sequence. Then we provide two videos (one is generated by our StableAnimator and the other is synthesized by a competitor) in random orders. Participants are then asked to answer the following questions: 
M-A/A-A/B-A/I-A: ``Which one has better motion/appearance/background/ID alignment with the reference". Table \ref{table:user_study} shows the superiority of our model regarding subjective evaluation.

\begin{table}[t!]\small
\caption{User preference of StableAnimator compared with other competitors. Higher indicates users prefer more to our model.
}
\vspace{-0.25in}
\begin{center}
\renewcommand\arraystretch{1.1}
\scalebox{0.9}{
\begin{tabular}{lcccc}
\toprule
Model         & M-A    & A-A    & B-A    & I-A    \\ \midrule
DisCo~\cite{wang2024disco}         & 95.6\% & 96.8\% & 94.2\% & 98.7\% \\
MagicAnimate~\cite{xu2024magicanimate}  & 94.5\% & 94.8\% & 92.7\% & 97.4\% \\
AnimateAnyone~\cite{hu2024animate} & 94.8\% & 93.1\% & 92.5\% & 98.3\% \\
Champ~\cite{zhu2024champ}         & 95.0\% & 91.3\% & 91.7\% & 96.6\% \\
Unianimate~\cite{wang2024unianimate}    & 89.7\% & 88.4\% & 90.5\% & 95.8\% \\
MimicMotion~\cite{zhang2024mimicmotion}   & 95.3\% & 95.5\% & 94.1\% & 97.6\% \\
ControlNeXt~\cite{peng2024controlnext}  & 93.6\% & 92.4\% & 90.3\% & 96.2\% \\ \bottomrule
\end{tabular}
}
\end{center}
\label{table:user_study}
\vspace{-0.3in}
\end{table}
\section{Conclusion}
\label{sec:conclusion}

In this paper, we proposed StableAnimator, a video diffusion model with dedicated modules for training and inference to generate high-quality, ID-preserving human image animations.
StableAnimator first used off-the-shelf models to gain image and face embeddings.
To capture the global context of the reference, StableAnimator introduced a Face Encoder to refine face embeddings.
StableAnimator further designed an ID-Adapter, which applied alignment to mitigate the interference from temporal modeling, enabling seamless face embedding integration without video fidelity loss.
During inference, to further enhance face quality, StableAnimator incorporated the HJB equation alongside diffusion denoising for face optimization.
It ran in parallel with denoising, creating an end-to-end pipeline that eliminates the need for third-party face-swapping tools.
Experimental results across various datasets demonstrated the superiority of our model in producing high-quality ID-preserving human animations.

StableAnimator is purely a research project. Currently, we have no plans to incorporate StableAnimator into a product or expand access to the public. In our research paper, we account for the ethical concerns associated with image-to-video research. To mitigate issues associated with training data, we have implemented a rigorous filtering process to purge our training data of inappropriate content, such as explicit imagery and offensive language, to minimize the likelihood of generating inappropriate content. 
{
    \small
    \bibliographystyle{ieeenat_fullname}
    \bibliography{main}

\begin{thebibliography}{66}
\providecommand{\natexlab}[1]{#1}
\providecommand{\url}[1]{\texttt{#1}}
\expandafter\ifx\csname urlstyle\endcsname\relax
  \providecommand{\doi}[1]{doi: #1}\else
  \providecommand{\doi}{doi: \begingroup \urlstyle{rm}\Url}\fi

\bibitem[Bao et~al.(2024)Bao, Xiang, Yue, He, Zhu, Zheng, Zhao, Liu, Wang, and Zhu]{bao2024vidu}
Fan Bao, Chendong Xiang, Gang Yue, Guande He, Hongzhou Zhu, Kaiwen Zheng, Min Zhao, Shilong Liu, Yaole Wang, and Jun Zhu.
\newblock Vidu: a highly consistent, dynamic and skilled text-to-video generator with diffusion models.
\newblock \emph{arXiv preprint arXiv:2405.04233}, 2024.

\bibitem[Bardi et~al.(1997)Bardi, Dolcetta, et~al.]{bardi1997optimal}
Martino Bardi, Italo~Capuzzo Dolcetta, et~al.
\newblock \emph{Optimal control and viscosity solutions of Hamilton-Jacobi-Bellman equations}.
\newblock Springer, 1997.

\bibitem[Blattmann et~al.(2023)Blattmann, Dockhorn, Kulal, Mendelevitch, Kilian, Lorenz, Levi, English, Voleti, Letts, et~al.]{blattmann2023stable}
Andreas Blattmann, Tim Dockhorn, Sumith Kulal, Daniel Mendelevitch, Maciej Kilian, Dominik Lorenz, Yam Levi, Zion English, Vikram Voleti, Adam Letts, et~al.
\newblock Stable video diffusion: Scaling latent video diffusion models to large datasets.
\newblock \emph{arXiv preprint arXiv:2311.15127}, 2023.

\bibitem[Brooks et~al.(2024)Brooks, Peebles, Holmes, DePue, Guo, Jing, Schnurr, Taylor, Luhman, Luhman, Ng, Wang, and Ramesh]{videoworldsimulators2024}
Tim Brooks, Bill Peebles, Connor Holmes, Will DePue, Yufei Guo, Li Jing, David Schnurr, Joe Taylor, Troy Luhman, Eric Luhman, Clarence Ng, Ricky Wang, and Aditya Ramesh.
\newblock Video generation models as world simulators.
\newblock 2024.

\bibitem[Chen et~al.(2024)Chen, Gu, Dinh, Theodorou, Susskind, and Zhai]{chen2023generative}
Tianrong Chen, Jiatao Gu, Laurent Dinh, Evangelos~A Theodorou, Joshua Susskind, and Shuangfei Zhai.
\newblock Generative modeling with phase stochastic bridges.
\newblock In \emph{ICLR}, 2024.

\bibitem[Dao and Gu(2024)]{mamba2}
Tri Dao and Albert Gu.
\newblock Transformers are {SSM}s: Generalized models and efficient algorithms through structured state space duality.
\newblock In \emph{ICML}, 2024.

\bibitem[Deng et~al.(2019)Deng, Guo, Xue, and Zafeiriou]{deng2019arcface}
Jiankang Deng, Jia Guo, Niannan Xue, and Stefanos Zafeiriou.
\newblock Arcface: Additive angular margin loss for deep face recognition.
\newblock In \emph{CVPR}, 2019.

\bibitem[Dhariwal and Nichol(2021)]{dhariwal2021diffusion}
Prafulla Dhariwal and Alexander Nichol.
\newblock Diffusion models beat gans on image synthesis.
\newblock In \emph{NeurIPS}, 2021.

\bibitem[Efron(2011)]{efron2011tweedie}
Bradley Efron.
\newblock Tweedie’s formula and selection bias.
\newblock \emph{Journal of the American Statistical Association}, 2011.

\bibitem[Fleming and Rishel(2012)]{fleming2012deterministic}
Wendell~H Fleming and Raymond~W Rishel.
\newblock \emph{Deterministic and stochastic optimal control}.
\newblock Springer Science \& Business Media, 2012.

\bibitem[Goodfellow et~al.(2020)Goodfellow, Pouget-Abadie, Mirza, Xu, Warde-Farley, Ozair, Courville, and Bengio]{goodfellow2020generative}
Ian Goodfellow, Jean Pouget-Abadie, Mehdi Mirza, Bing Xu, David Warde-Farley, Sherjil Ozair, Aaron Courville, and Yoshua Bengio.
\newblock Generative adversarial networks.
\newblock \emph{Communications of the ACM}, 2020.

\bibitem[Guo et~al.(2024{\natexlab{a}})Guo, Zhang, Liu, Zhong, Zhang, Wan, and Zhang]{guo2024liveportrait}
Jianzhu Guo, Dingyun Zhang, Xiaoqiang Liu, Zhizhou Zhong, Yuan Zhang, Pengfei Wan, and Di Zhang.
\newblock Liveportrait: Efficient portrait animation with stitching and retargeting control.
\newblock \emph{arXiv preprint arXiv:2407.03168}, 2024{\natexlab{a}}.

\bibitem[Guo et~al.(2024{\natexlab{b}})Guo, Yang, Rao, Liang, Wang, Qiao, Agrawala, Lin, and Dai]{guo2023animatediff}
Yuwei Guo, Ceyuan Yang, Anyi Rao, Zhengyang Liang, Yaohui Wang, Yu Qiao, Maneesh Agrawala, Dahua Lin, and Bo Dai.
\newblock Animatediff: Animate your personalized text-to-image diffusion models without specific tuning.
\newblock In \emph{ICLR}, 2024{\natexlab{b}}.

\bibitem[Guo et~al.(2024{\natexlab{c}})Guo, Wu, Chen, Chen, and He]{guo2024pulid}
Zinan Guo, Yanze Wu, Zhuowei Chen, Lang Chen, and Qian He.
\newblock Pulid: Pure and lightning id customization via contrastive alignment.
\newblock In \emph{NeurIPS}, 2024{\natexlab{c}}.

\bibitem[Henry(2024)]{facefusion}
Ruhs Henry.
\newblock Facefusion.
\newblock \url{https://github.com/facefusion/facefusion}, 2024.

\bibitem[Hertz et~al.(2022)Hertz, Mokady, Tenenbaum, Aberman, Pritch, and Cohen-Or]{hertz2022prompt}
Amir Hertz, Ron Mokady, Jay Tenenbaum, Kfir Aberman, Yael Pritch, and Daniel Cohen-Or.
\newblock Prompt-to-prompt image editing with cross attention control.
\newblock \emph{arXiv preprint arXiv:2208.01626}, 2022.

\bibitem[Ho et~al.(2020)Ho, Jain, and Abbeel]{ho2020denoising}
Jonathan Ho, Ajay Jain, and Pieter Abbeel.
\newblock Denoising diffusion probabilistic models.
\newblock In \emph{NeurIPS}, 2020.

\bibitem[Ho et~al.(2022)Ho, Saharia, Chan, Fleet, Norouzi, and Salimans]{ho2022cascaded}
Jonathan Ho, Chitwan Saharia, William Chan, David~J Fleet, Mohammad Norouzi, and Tim Salimans.
\newblock Cascaded diffusion models for high fidelity image generation.
\newblock \emph{JMLR}, 2022.

\bibitem[Hong et~al.(2022)Hong, Ding, Zheng, Liu, and Tang]{hong2022cogvideo}
Wenyi Hong, Ming Ding, Wendi Zheng, Xinghan Liu, and Jie Tang.
\newblock Cogvideo: Large-scale pretraining for text-to-video generation via transformers.
\newblock \emph{arXiv preprint arXiv:2205.15868}, 2022.

\bibitem[Hore and Ziou(2010)]{hore2010image}
Alain Hore and Djemel Ziou.
\newblock Image quality metrics: Psnr vs. ssim.
\newblock In \emph{2010 20th international conference on pattern recognition}, 2010.

\bibitem[Hu et~al.(2021)Hu, Shen, Wallis, Allen-Zhu, Li, Wang, Wang, and Chen]{hu2021lora}
Edward~J Hu, Yelong Shen, Phillip Wallis, Zeyuan Allen-Zhu, Yuanzhi Li, Shean Wang, Lu Wang, and Weizhu Chen.
\newblock Lora: Low-rank adaptation of large language models.
\newblock In \emph{ICLR}, 2021.

\bibitem[Hu(2024)]{hu2024animate}
Li Hu.
\newblock Animate anyone: Consistent and controllable image-to-video synthesis for character animation.
\newblock In \emph{CVPR}, 2024.

\bibitem[Huang et~al.(2024)Huang, Dong, Song, Li, Zhou, Cheng, Liao, Chen, Yan, Liao, et~al.]{huang2024consistentid}
Jiehui Huang, Xiao Dong, Wenhui Song, Hanhui Li, Jun Zhou, Yuhao Cheng, Shutao Liao, Long Chen, Yiqiang Yan, Shengcai Liao, et~al.
\newblock Consistentid: Portrait generation with multimodal fine-grained identity preserving.
\newblock \emph{arXiv preprint arXiv:2404.16771}, 2024.

\bibitem[Huang et~al.(2021)Huang, Han, Xu, and Zhang]{huang2021few}
Zhichao Huang, Xintong Han, Jia Xu, and Tong Zhang.
\newblock Few-shot human motion transfer by personalized geometry and texture modeling.
\newblock In \emph{CVPR}, 2021.

\bibitem[Jafarian and Park(2021)]{jafarian2021learning}
Yasamin Jafarian and Hyun~Soo Park.
\newblock Learning high fidelity depths of dressed humans by watching social media dance videos.
\newblock In \emph{CVPR}, 2021.

\bibitem[Karras et~al.(2022)Karras, Aittala, Aila, and Laine]{karras2022elucidating}
Tero Karras, Miika Aittala, Timo Aila, and Samuli Laine.
\newblock Elucidating the design space of diffusion-based generative models.
\newblock In \emph{NeurIPS}, 2022.

\bibitem[Kingma(2014)]{kingma2013auto}
Diederik~P Kingma.
\newblock Auto-encoding variational bayes.
\newblock In \emph{ICLR}, 2014.

\bibitem[Kirk(2004)]{kirk2004optimal}
Donald~E Kirk.
\newblock \emph{Optimal control theory: an introduction}.
\newblock Courier Corporation, 2004.

\bibitem[Li et~al.(2024)Li, Cao, Wang, Qi, Cheng, and Shan]{li2024photomaker}
Zhen Li, Mingdeng Cao, Xintao Wang, Zhongang Qi, Ming-Ming Cheng, and Ying Shan.
\newblock Photomaker: Customizing realistic human photos via stacked id embedding.
\newblock In \emph{CVPR}, 2024.

\bibitem[Ma et~al.(2024)Ma, Wang, Jia, Chen, Liu, Li, Chen, and Qiao]{ma2024latte}
Xin Ma, Yaohui Wang, Gengyun Jia, Xinyuan Chen, Ziwei Liu, Yuan-Fang Li, Cunjian Chen, and Yu Qiao.
\newblock Latte: Latent diffusion transformer for video generation.
\newblock \emph{arXiv preprint arXiv:2401.03048}, 2024.

\bibitem[Meng et~al.(2021)Meng, He, Song, Song, Wu, Zhu, and Ermon]{meng2021sdedit}
Chenlin Meng, Yutong He, Yang Song, Jiaming Song, Jiajun Wu, Jun-Yan Zhu, and Stefano Ermon.
\newblock Sdedit: Guided image synthesis and editing with stochastic differential equations.
\newblock In \emph{ICLR}, 2021.

\bibitem[Nichol and Dhariwal(2021)]{nichol2021improved}
Alexander~Quinn Nichol and Prafulla Dhariwal.
\newblock Improved denoising diffusion probabilistic models.
\newblock In \emph{ICML}, 2021.

\bibitem[Peebles and Xie(2023)]{peebles2023scalable}
William Peebles and Saining Xie.
\newblock Scalable diffusion models with transformers.
\newblock In \emph{ICCV}, 2023.

\bibitem[Peng et~al.(2024)Peng, Wang, Zhang, Li, Yang, and Jia]{peng2024controlnext}
Bohao Peng, Jian Wang, Yuechen Zhang, Wenbo Li, Ming-Chang Yang, and Jiaya Jia.
\newblock Controlnext: Powerful and efficient control for image and video generation.
\newblock \emph{arXiv preprint arXiv:2408.06070}, 2024.

\bibitem[Peng(1992)]{peng1992stochastic}
Shige Peng.
\newblock Stochastic hamilton--jacobi--bellman equations.
\newblock \emph{SIAM Journal on Control and Optimization}, 1992.

\bibitem[Radford et~al.(2021)Radford, Kim, Hallacy, Ramesh, Goh, Agarwal, Sastry, Askell, Mishkin, Clark, et~al.]{radford2021learning}
Alec Radford, Jong~Wook Kim, Chris Hallacy, Aditya Ramesh, Gabriel Goh, Sandhini Agarwal, Girish Sastry, Amanda Askell, Pamela Mishkin, Jack Clark, et~al.
\newblock Learning transferable visual models from natural language supervision.
\newblock In \emph{ICML}, 2021.

\bibitem[Rombach et~al.(2022)Rombach, Blattmann, Lorenz, Esser, and Ommer]{rombach2022high}
Robin Rombach, Andreas Blattmann, Dominik Lorenz, Patrick Esser, and Bj{\"o}rn Ommer.
\newblock High-resolution image synthesis with latent diffusion models.
\newblock In \emph{CVPR}, 2022.

\bibitem[Siarohin et~al.(2019)Siarohin, Lathuilière, Tulyakov, Ricci, and Sebe]{Siarohin_2019_NeurIPS}
Aliaksandr Siarohin, Stéphane Lathuilière, Sergey Tulyakov, Elisa Ricci, and Nicu Sebe.
\newblock First order motion model for image animation.
\newblock In \emph{NeurIPS}, 2019.

\bibitem[Siarohin et~al.(2021)Siarohin, Woodford, Ren, Chai, and Tulyakov]{siarohin2021motion}
Aliaksandr Siarohin, Oliver~J Woodford, Jian Ren, Menglei Chai, and Sergey Tulyakov.
\newblock Motion representations for articulated animation.
\newblock In \emph{CVPR}, 2021.

\bibitem[Singer et~al.(2022)Singer, Polyak, Hayes, Yin, An, Zhang, Hu, Yang, Ashual, Gafni, et~al.]{singer2022make}
Uriel Singer, Adam Polyak, Thomas Hayes, Xi Yin, Jie An, Songyang Zhang, Qiyuan Hu, Harry Yang, Oron Ashual, Oran Gafni, et~al.
\newblock Make-a-video: Text-to-video generation without text-video data.
\newblock \emph{arXiv preprint arXiv:2209.14792}, 2022.

\bibitem[Song et~al.(2021{\natexlab{a}})Song, Meng, and Ermon]{song2020denoising}
Jiaming Song, Chenlin Meng, and Stefano Ermon.
\newblock Denoising diffusion implicit models.
\newblock In \emph{ICLR}, 2021{\natexlab{a}}.

\bibitem[Song et~al.(2021{\natexlab{b}})Song, Sohl-Dickstein, Kingma, Kumar, Ermon, and Poole]{song2020score}
Yang Song, Jascha Sohl-Dickstein, Diederik~P Kingma, Abhishek Kumar, Stefano Ermon, and Ben Poole.
\newblock Score-based generative modeling through stochastic differential equations.
\newblock In \emph{ICLR}, 2021{\natexlab{b}}.

\bibitem[Tu et~al.(2024{\natexlab{a}})Tu, Dai, Cheng, Hu, Han, Wu, and Jiang]{tu2024motioneditor}
Shuyuan Tu, Qi Dai, Zhi-Qi Cheng, Han Hu, Xintong Han, Zuxuan Wu, and Yu-Gang Jiang.
\newblock Motioneditor: Editing video motion via content-aware diffusion.
\newblock In \emph{CVPR}, 2024{\natexlab{a}}.

\bibitem[Tu et~al.(2024{\natexlab{b}})Tu, Dai, Zhang, Xie, Cheng, Luo, Han, Wu, and Jiang]{tu2024motionfollower}
Shuyuan Tu, Qi Dai, Zihao Zhang, Sicheng Xie, Zhi-Qi Cheng, Chong Luo, Xintong Han, Zuxuan Wu, and Yu-Gang Jiang.
\newblock Motionfollower: Editing video motion via lightweight score-guided diffusion.
\newblock \emph{arXiv preprint arXiv:2405.20325}, 2024{\natexlab{b}}.

\bibitem[Tumanyan et~al.(2023)Tumanyan, Geyer, Bagon, and Dekel]{tumanyan2023plug}
Narek Tumanyan, Michal Geyer, Shai Bagon, and Tali Dekel.
\newblock Plug-and-play diffusion features for text-driven image-to-image translation.
\newblock In \emph{CVPR}, 2023.

\bibitem[Wang et~al.(2024{\natexlab{a}})Wang, Bai, Wang, Qin, and Chen]{wang2024instantid}
Qixun Wang, Xu Bai, Haofan Wang, Zekui Qin, and Anthony Chen.
\newblock Instantid: Zero-shot identity-preserving generation in seconds.
\newblock \emph{arXiv preprint arXiv:2401.07519}, 2024{\natexlab{a}}.

\bibitem[Wang et~al.(2024{\natexlab{b}})Wang, Li, Lin, Zhai, Lin, Yang, Zhang, Liu, and Wang]{wang2024disco}
Tan Wang, Linjie Li, Kevin Lin, Yuanhao Zhai, Chung-Ching Lin, Zhengyuan Yang, Hanwang Zhang, Zicheng Liu, and Lijuan Wang.
\newblock Disco: Disentangled control for realistic human dance generation.
\newblock In \emph{CVPR}, 2024{\natexlab{b}}.

\bibitem[Wang et~al.(2024{\natexlab{c}})Wang, Liu, Lin, Yan, Chen, Low, Hoang, Wu, Liew, Yan, et~al.]{wang2024magicvideo}
Weimin Wang, Jiawei Liu, Zhijie Lin, Jiangqiao Yan, Shuo Chen, Chetwin Low, Tuyen Hoang, Jie Wu, Jun~Hao Liew, Hanshu Yan, et~al.
\newblock Magicvideo-v2: Multi-stage high-aesthetic video generation.
\newblock \emph{arXiv preprint arXiv:2401.04468}, 2024{\natexlab{c}}.

\bibitem[Wang et~al.(2021)Wang, Li, Zhang, and Shan]{wang2021gfpgan}
Xintao Wang, Yu Li, Honglun Zhang, and Ying Shan.
\newblock Towards real-world blind face restoration with generative facial prior.
\newblock In \emph{The IEEE Conference on Computer Vision and Pattern Recognition (CVPR)}, 2021.

\bibitem[Wang et~al.(2024{\natexlab{d}})Wang, Zhang, Gao, Wang, Zhou, Zhang, Yan, and Sang]{wang2024unianimate}
Xiang Wang, Shiwei Zhang, Changxin Gao, Jiayu Wang, Xiaoqiang Zhou, Yingya Zhang, Luxin Yan, and Nong Sang.
\newblock Unianimate: Taming unified video diffusion models for consistent human image animation.
\newblock \emph{arXiv preprint arXiv:2406.01188}, 2024{\natexlab{d}}.

\bibitem[Weng et~al.(2024)Weng, Yang, Xing, Wu, and Jiang]{weng2024genrec}
Zejia Weng, Xitong Yang, Zhen Xing, Zuxuan Wu, and Yu-Gang Jiang.
\newblock Genrec: Unifying video generation and recognition with diffusion models.
\newblock \emph{arXiv preprint arXiv:2408.15241}, 2024.

\bibitem[Wu et~al.(2023)Wu, Ge, Wang, Lei, Gu, Shi, Hsu, Shan, Qie, and Shou]{wu2023tune}
Jay~Zhangjie Wu, Yixiao Ge, Xintao Wang, Stan~Weixian Lei, Yuchao Gu, Yufei Shi, Wynne Hsu, Ying Shan, Xiaohu Qie, and Mike~Zheng Shou.
\newblock Tune-a-video: One-shot tuning of image diffusion models for text-to-video generation.
\newblock In \emph{CVPR}, 2023.

\bibitem[Xing et~al.(2023)Xing, Dai, Zhang, Zhang, Hu, Wu, and Jiang]{xing2023vidiff}
Zhen Xing, Qi Dai, Zihao Zhang, Hui Zhang, Han Hu, Zuxuan Wu, and Yu-Gang Jiang.
\newblock Vidiff: Translating videos via multi-modal instructions with diffusion models.
\newblock \emph{arXiv preprint arXiv:2311.18837}, 2023.

\bibitem[Xing et~al.(2024{\natexlab{a}})Xing, Dai, Hu, Wu, and Jiang]{xing2024simda}
Zhen Xing, Qi Dai, Han Hu, Zuxuan Wu, and Yu-Gang Jiang.
\newblock Simda: Simple diffusion adapter for efficient video generation.
\newblock In \emph{Proceedings of the IEEE/CVF Conference on Computer Vision and Pattern Recognition}, pages 7827--7839, 2024{\natexlab{a}}.

\bibitem[Xing et~al.(2024{\natexlab{b}})Xing, Dai, Weng, Wu, and Jiang]{xing2024aid}
Zhen Xing, Qi Dai, Zejia Weng, Zuxuan Wu, and Yu-Gang Jiang.
\newblock Aid: Adapting image2video diffusion models for instruction-guided video prediction.
\newblock \emph{arXiv preprint arXiv:2406.06465}, 2024{\natexlab{b}}.

\bibitem[Xing et~al.(2024{\natexlab{c}})Xing, Feng, Chen, Dai, Hu, Xu, Wu, and Jiang]{xing2024survey}
Zhen Xing, Qijun Feng, Haoran Chen, Qi Dai, Han Hu, Hang Xu, Zuxuan Wu, and Yu-Gang Jiang.
\newblock A survey on video diffusion models.
\newblock \emph{ACM Computing Surveys}, 57\penalty0 (2):\penalty0 1--42, 2024{\natexlab{c}}.

\bibitem[Xu et~al.(2024)Xu, Zhang, Liew, Yan, Liu, Zhang, Feng, and Shou]{xu2024magicanimate}
Zhongcong Xu, Jianfeng Zhang, Jun~Hao Liew, Hanshu Yan, Jia-Wei Liu, Chenxu Zhang, Jiashi Feng, and Mike~Zheng Shou.
\newblock Magicanimate: Temporally consistent human image animation using diffusion model.
\newblock In \emph{CVPR}, 2024.

\bibitem[Yan et~al.(2021)Yan, Zhang, Abbeel, and Srinivas]{yan2021videogpt}
Wilson Yan, Yunzhi Zhang, Pieter Abbeel, and Aravind Srinivas.
\newblock Videogpt: Video generation using vq-vae and transformers.
\newblock \emph{arXiv preprint arXiv:2104.10157}, 2021.

\bibitem[Yan et~al.(2023)Yan, Zhang, Wang, Zhou, Zhang, Cheng, Yu, and Fu]{yan2023facestudio}
Yuxuan Yan, Chi Zhang, Rui Wang, Yichao Zhou, Gege Zhang, Pei Cheng, Gang Yu, and Bin Fu.
\newblock Facestudio: Put your face everywhere in seconds.
\newblock \emph{arXiv preprint arXiv:2312.02663}, 2023.

\bibitem[Yang et~al.(2023)Yang, Zeng, Yuan, and Li]{yang2023effective}
Zhendong Yang, Ailing Zeng, Chun Yuan, and Yu Li.
\newblock Effective whole-body pose estimation with two-stages distillation.
\newblock In \emph{ICCV}, 2023.

\bibitem[Ye et~al.(2023)Ye, Zhang, Liu, Han, and Yang]{ye2023ip-adapter}
Hu Ye, Jun Zhang, Sibo Liu, Xiao Han, and Wei Yang.
\newblock Ip-adapter: Text compatible image prompt adapter for text-to-image diffusion models.
\newblock \emph{arXiv preprint arxiv:2308.06721}, 2023.

\bibitem[Yu et~al.(2023)Yu, Cheng, Sohn, Lezama, Zhang, Chang, Hauptmann, Yang, Hao, Essa, et~al.]{yu2023magvit}
Lijun Yu, Yong Cheng, Kihyuk Sohn, Jos{\'e} Lezama, Han Zhang, Huiwen Chang, Alexander~G Hauptmann, Ming-Hsuan Yang, Yuan Hao, Irfan Essa, et~al.
\newblock Magvit: Masked generative video transformer.
\newblock In \emph{CVPR}, pages 10459--10469, 2023.

\bibitem[Zhang et~al.(2023)Zhang, Rao, and Agrawala]{zhang2023adding}
Lvmin Zhang, Anyi Rao, and Maneesh Agrawala.
\newblock Adding conditional control to text-to-image diffusion models.
\newblock In \emph{ICCV}, 2023.

\bibitem[Zhang et~al.(2024)Zhang, Gu, Wang, Wang, Cheng, Zhu, and Zou]{zhang2024mimicmotion}
Yuang Zhang, Jiaxi Gu, Li-Wen Wang, Han Wang, Junqi Cheng, Yuefeng Zhu, and Fangyuan Zou.
\newblock Mimicmotion: High-quality human motion video generation with confidence-aware pose guidance.
\newblock \emph{arXiv preprint arXiv:2406.19680}, 2024.

\bibitem[Zhou et~al.(2022)Zhou, Chan, Li, and Loy]{zhou2022codeformer}
Shangchen Zhou, Kelvin~C.K. Chan, Chongyi Li, and Chen~Change Loy.
\newblock Towards robust blind face restoration with codebook lookup transformer.
\newblock In \emph{NeurIPS}, 2022.

\bibitem[Zhu et~al.(2024)Zhu, Chen, Dai, Xu, Cao, Yao, Zhu, and Zhu]{zhu2024champ}
Shenhao Zhu, Junming~Leo Chen, Zuozhuo Dai, Yinghui Xu, Xun Cao, Yao Yao, Hao Zhu, and Siyu Zhu.
\newblock Champ: Controllable and consistent human image animation with 3d parametric guidance.
\newblock In \emph{EECV}, 2024.

\end{thebibliography}
}
\clearpage
\setcounter{page}{1}
\appendix
\section{Supplementary Material}

\subsection{Evaluation Metrics}
Following previous human image animation evaluation settings, we implement numerous quantitative evaluation metrics, including L1, PSNR, SSIM, LPIPS, FVD, and CSIM, to compare our StableAnimator with current state-of-the-art animation models.
The details of the above metrics are described as follows:
\begin{enumerate}[label=(\arabic*)]
    \item L1 refers to the average absolute difference between the corresponding pixel values of two images. It measures the typical magnitude of prediction errors without considering their direction, making it a valuable tool for quantifying the extent of discrepancies.
    \item PSNR measures the ratio between the maximum possible power of a signal (in this case, the original image) and the power of corrupting noise that affects the fidelity of its representation. PSNR is expressed in decibels (dB), with higher values indicating better quality.
    \item SSIM refers to the similarity between two images based on their luminance, contrast, and structural information. 
    \item LPIPS measures the similarity between images by analyzing the feature representations of their patches, reflecting human visual perception effectively.
    \item FVD evaluates the disparity between the feature distributions of real and generated videos, considering both spatial and temporal dimensions. FVD is often used to measure the video fidelity.
    \item CSIM refers to the cosine similarity between the facial embeddings of two face images. The facial embeddings are extracted by ArcFace.
\end{enumerate}

\subsection{Preliminaries}
The diffusion model includes a forward diffusion process and a reverse denoising process. In the forward process, the Gaussian noise is progressively added to the data sample $\bm{x}_{0}\sim\bm{p}_{\text{data}}$ from the particular data distribution $\bm{p}_{\text{data}}$:
\begin{equation}\small
\label{eq:forward_diffusion}
\begin{aligned}
     \bm{q}(\bm{x}_{t}|\bm{x}_{t-1}) = \mathcal{N}(\bm{x}_{t}; \sqrt{\alpha_{t}}\bm{x}_{t-1},(1-\alpha_{t})\mathbf{I}).
\end{aligned}
\end{equation}
The data sample $\bm{x}_{0}$ is ultimately converted into Gaussian noise $\bm{x}_{T}\sim\mathcal{N}(0,1)$ after $\bm{T}$ diffusion forward steps. $\alpha_{t}$ is a constant noise schedule. In the reverse process, the diffusion model $\bm{\varepsilon}_{\theta}(\bm{x}_{t},t)$ tends to recover $\bm{x}_{0}$ from $\bm{x}_{T}$ by predicting the noise $\bm{\varepsilon}$ based on the current sample $\bm{x}_{t}$ and time step $\bm{t}$. The MSE loss is applied to train $\bm{\varepsilon}(\cdot)$:
\begin{equation}\small
\label{eq:mse_loss}
\begin{aligned}
     \mathcal{L} = \mathbb{E}_{\bm{x}_{0},\bm{\varepsilon},t}(\left \| \bm{\varepsilon} -\bm{\varepsilon}_{\theta}(\bm{x}_{t}, t)  \right \|^{2}).
\end{aligned}
\end{equation}
Moreover, the denoising process can be regarded as a continuous process (reverse-SDE):
\begin{equation}\small
\label{eq:sde}
\begin{aligned}
     d\bm{X}_{t}=[f(\bm{X}_{t}, \bm{t})-g^{2}(\bm{X}_{t}, \bm{t})\nabla \log p(\bm{X}_{t}, \bm{t})]d\bm{t}+g(\bm{X}_{t}, \bm{t})d\bm{W}_{t},
\end{aligned}
\end{equation}
where $\bm{W}_{t}$ and $\nabla \log p(\bm{X}_{t}, \bm{t})$ refer to the standard Brownian motion and score function. $f(\bm{X}_{t}, \bm{t})$ and $g(\bm{X}_{t}, \bm{t})$ are drift and volatility. The diffusion model $\bm{\varepsilon}_{\theta}(\bm{x}_{t},t)$ approximates $\nabla \log p(\bm{X}_{t}, \bm{t})$ during the continuous denoising process.

\begin{figure}[t!]
\begin{center}
\includegraphics[width=1\linewidth]{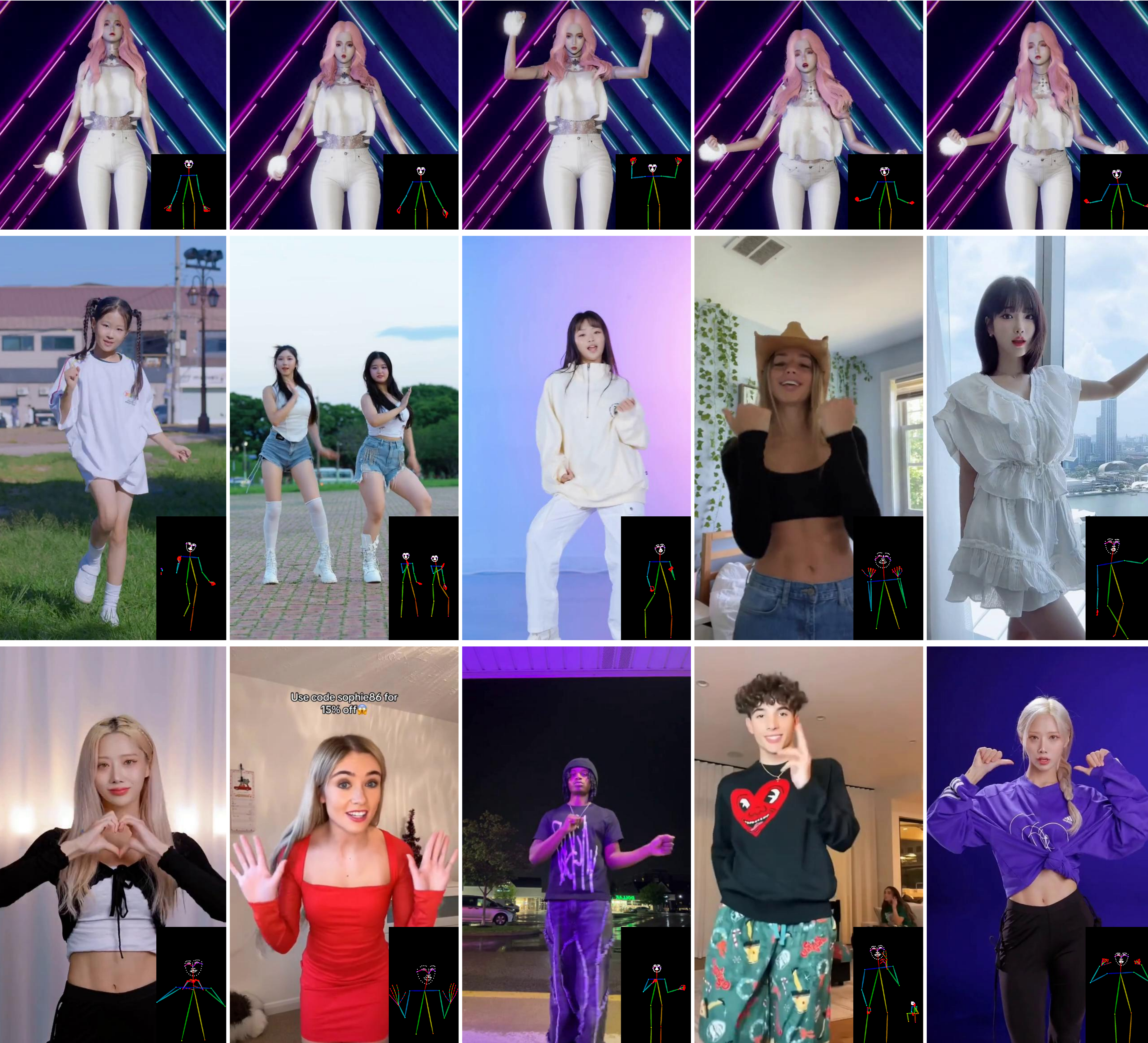}
\end{center}
\vspace{-0.5cm}
   \caption{Examples from Unseen100.
   }
\label{fig:unseen100}
\vspace{-0.5cm}
\end{figure}

\begin{algorithm*}[t!]
\caption{HJB Equation-based Face Optimization ($\sigma(t) = t$ and $s(t) = 1$)}
\label{alg:face_optimization}
\begin{algorithmic}
\State \textbf{Input:} {$\text{A diffusion model }\mathtt{D}_{\theta}(\bm{x}; \bm{\sigma}),\text{Timesteps }t_{i\in \{0, \ldots, N\}}, \text{Pre-defined factors }\bm{\gamma}_{i\in \{0, \ldots, N-1\}}, \text{A reference image }\bm{y}$} 
    \State \textbf{Sample} $\bm{x}_0 \sim \mathcal{N}(0, t_0^2\bm{I})$
    \State \textbf{For} $i \in \{0, \ldots, N-1\}$ \textbf{do} 
        \State \hspace{1em} $\bm{\gamma}_i = 0$ 
        \State \hspace{1em} \textbf{if} $t_i \in [\bm{S}_{t_{\text{min}}}, \bm{S}_{t_{\text{max}}}]:$ \hfill $\triangleright$ $\bm{S}_{\text{noise}}$,$ \bm{S}_{\text{churn}}$, $\bm{S}_{t_{\text{min}}}$, and $\bm{S}_{t_{\text{max}}}$ are the pre-defined values of EDM 
        \State \hspace{1em} \hspace{1em} $\bm{\gamma}_i = \min \left( \frac{\bm{S}_{\text{churn}}}{N}, \sqrt{2}-1 \right)$
        \State \hspace{1em} \textbf{Sample} $\bm{\epsilon}_i \sim \mathcal{N}(0, \bm{S}_{\text{noise}}^2\bm{I})$
        \State \hspace{1em} $\hat{t}_i = t_i + \bm{\gamma}_i t_i$ \hfill $\triangleright$ Select temporarily increased noise level $\hat{t}_i$
        \State \hspace{1em} $\hat{\bm{x}}_i = \bm{x}_i + \sqrt{\hat{t}_i^2 - t_i^2} \bm{\epsilon}_i$ \hfill $\triangleright$ Add new noise to move from $t_i$ to $\hat{t}_i$
        \State \hspace{1em} $\bm{x}_{\text{pred}}=\mathtt{D}_{\theta}(\hat{\bm{x}}_i; \hat{t}_i)$ \hfill $\triangleright$ The diffusion model predicts the denoised sample 
        \State \hspace{1em} $\bm{x}_{\text{op}}=\bm{x}_{\text{pred}}.\mathtt{clone}().\mathtt{detach}()$ \hfill $\triangleright$ Starting optimization 
        \State \hspace{1em} $\bm{op}=\mathtt{Adam}([\bm{x}_{\text{op}}], \bm{\eta})$ \hfill $\triangleright$ $\mathtt{Adam}$ and $\bm{\eta}$ are an Adam optimizer and a learning rate
        \State \hspace{1em} $\bm{x}_{\text{op}}.\text{requires\_grad}=\text{True}$ \hfill $\triangleright$ $\bm{x}_{\text{op}}$ is a HJB variable (trainable)
        \State \hspace{1em} \textbf{For} $k \in \{1,2, \ldots, 10\}$ \textbf{do} \hfill $\triangleright$ $k$ is the optimization step
        \State \hspace{1em} \hspace{1em} $\bm{f}_{\text{pred}}=\mathtt{Decoder}(\bm{x}_{\text{op}})$ \hfill $\triangleright$ $\mathtt{Decoder}$ is a VAE decoder, which converts predicted sample to the pixel level
        \State \hspace{1em} \hspace{1em} $\bm{loss}=(1-\mathtt{Cos}(\mathtt{Arc}(\bm{f}_{\text{pred}}), \mathtt{Arc}(\bm{y}))).\text{abs}().\text{mean}()$ \hfill $\triangleright$ $\mathtt{Cos}(\cdot)$ computes the similarity between given embeddings
        \State \hspace{1em} \hspace{1em}  $\bm{op}.\text{zero\_grad}()$ \hfill $\triangleright$ $\mathtt{Arc}$ is the Arcface model which extracts face embeddings
        \State \hspace{1em} \hspace{1em}  $\bm{loss}.\text{backward}(\text{retain\_graph=True})$ \hfill $\triangleright$ $\bm{x}_{\text{op}}$ is updated towards optimal face consistency by the gradient of the loss
        \State \hspace{1em} \hspace{1em}  $\bm{op}.\text{step}()$
        \State \hspace{1em} $\bm{x}_{\text{pred}}=\bm{x}_{\text{op}}$ \hfill $\triangleright$ End of Optimization 
        \State \hspace{1em} $\bm{d}_i = (\hat{\bm{x}}_i-\bm{x}_{\text{pred}})/\hat{t}_i$ \hfill $\triangleright$ Evaluate $d\bm{x}/dt$ at $\hat{t}_i$
        \State \hspace{1em} $\bm{x}_{i+1} = \hat{\bm{x}}_i + (t_{i+1} - \hat{t}_i)\bm{d}_i$ \hfill $\triangleright$ Take Euler step from $\hat{t}_i$ to $t_{i+1}$
        \State \hspace{1em} \textbf{if} $t_{i+1} \neq 0$:
        \State \hspace{1em} \hspace{1em} $\bm{d}'_i = (\bm{x}_{i+1}-\mathtt{D}_{\theta}(\bm{x}_{i+1}; t_{i+1}))/t_{i+1}$ \hfill $\triangleright$ Apply $2^{\text{nd}}$ order correction
        \State \hspace{1em} \hspace{1em} $\bm{x}_{i+1} = \hat{\bm{x}}_i + (t_{i+1} - \hat{t}_i) \left( \frac{1}{2} \bm{d}_i + \frac{1}{2} \bm{d}'_i \right)$
    \State \textbf{return} $\bm{x}_N$
\end{algorithmic}
\end{algorithm*}

\subsection{Details of Testing Dataset}
We select 100 unseen videos (10-20 seconds long) from the internet to construct the testing dataset Unseen100. 
Some examples are shown in Fig. \ref{fig:unseen100}.
The first row refers to five frames of a video, while the following rows represent individual frames of different videos.
The sources of videos come from numerous social media platforms, including YouTube, TikTok, and BiliBili. These videos showcase individuals across ethnicities, genders, portrayed in full-body, half-body, and close-up shots against varied indoor and outdoor settings. 
In contrast to existing open-source testing datasets (TikTok dataset), our Unseen100 contains relatively complicated motion information and intricate protagonist appearances. Moreover, positions and facial expressions in some Unseen100 videos dynamically change, such as shaking heads, making it more challenging to maintain identity consistency.

\subsection{Long Animation}
\label{long_animation}
\begin{figure*}[t!]
\begin{center}
\includegraphics[width=1\linewidth]{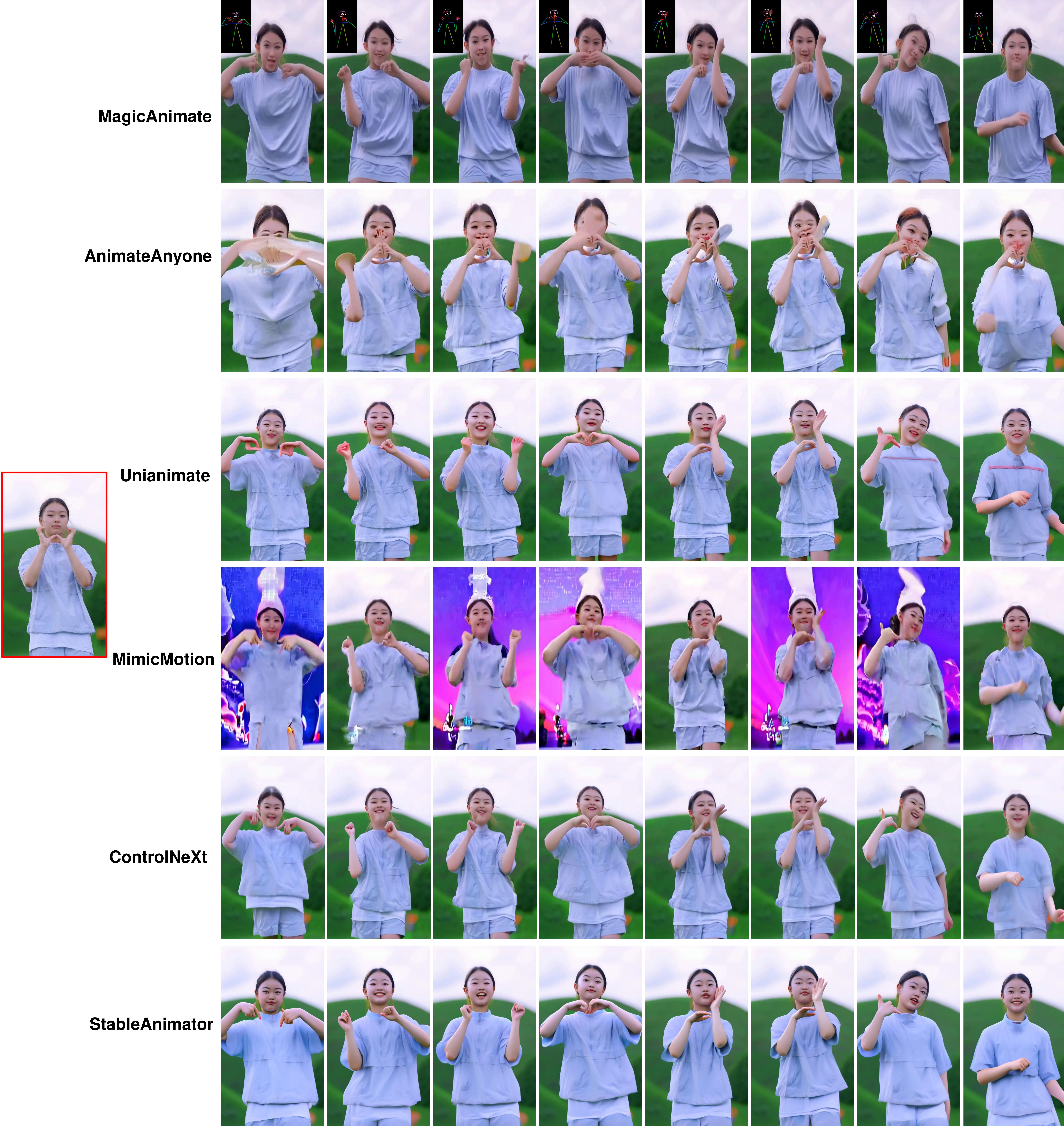}
\end{center}
\vspace{-0.55cm}
   \caption{Long animation results (1/3). The images with red borders are the reference images.}
\label{fig:long_video_1}
\vspace{-0.5cm}
\end{figure*}

\begin{figure*}[t!]
\begin{center}
\includegraphics[width=1\linewidth]{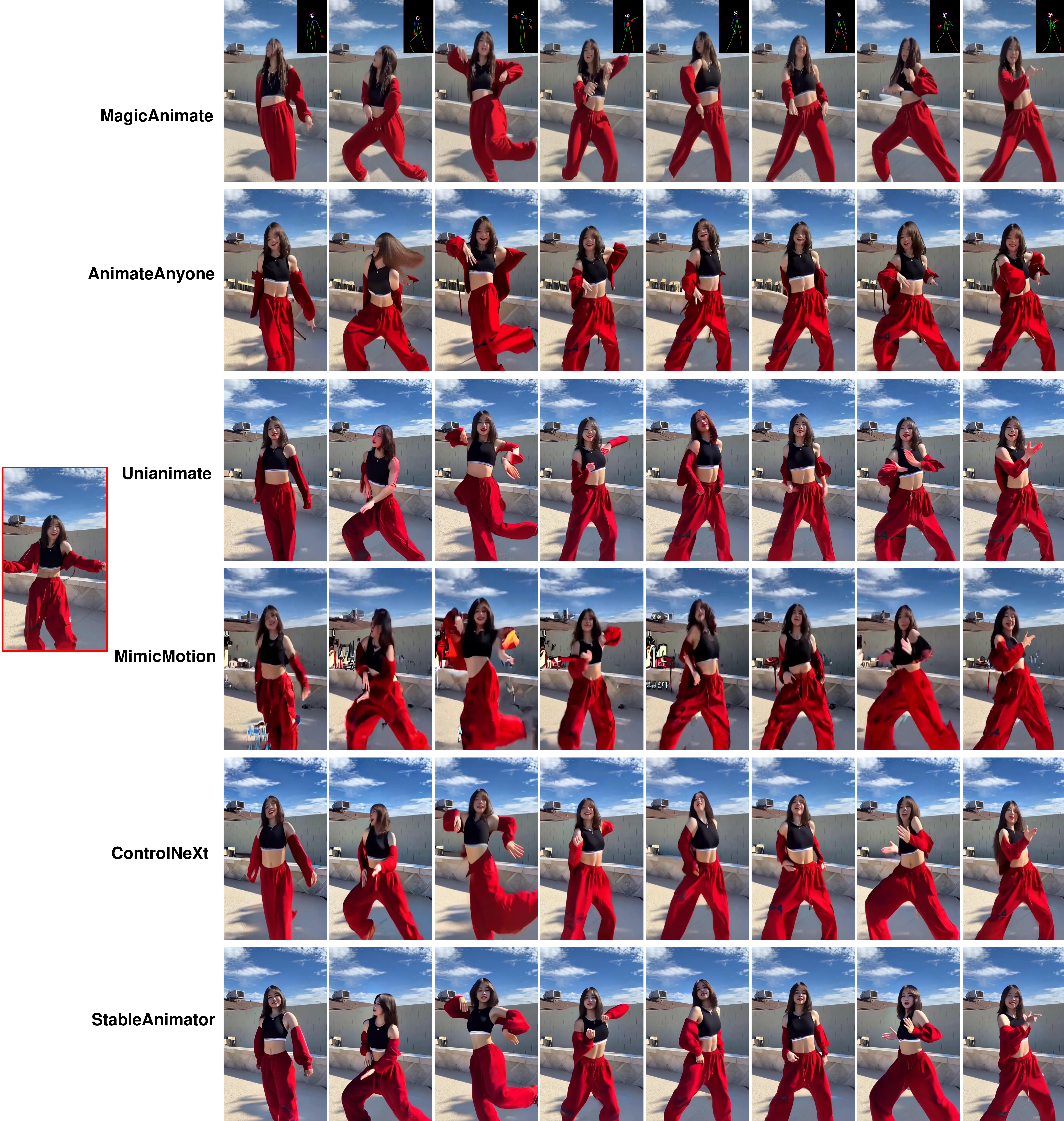}
\end{center}
\vspace{-0.55cm}
   \caption{Long animation results (2/3). The images with red borders are the reference images.}
\label{fig:long_video_2}
\vspace{-0.5cm}
\end{figure*}

\begin{figure*}[t!]
\begin{center}
\includegraphics[width=1\linewidth]{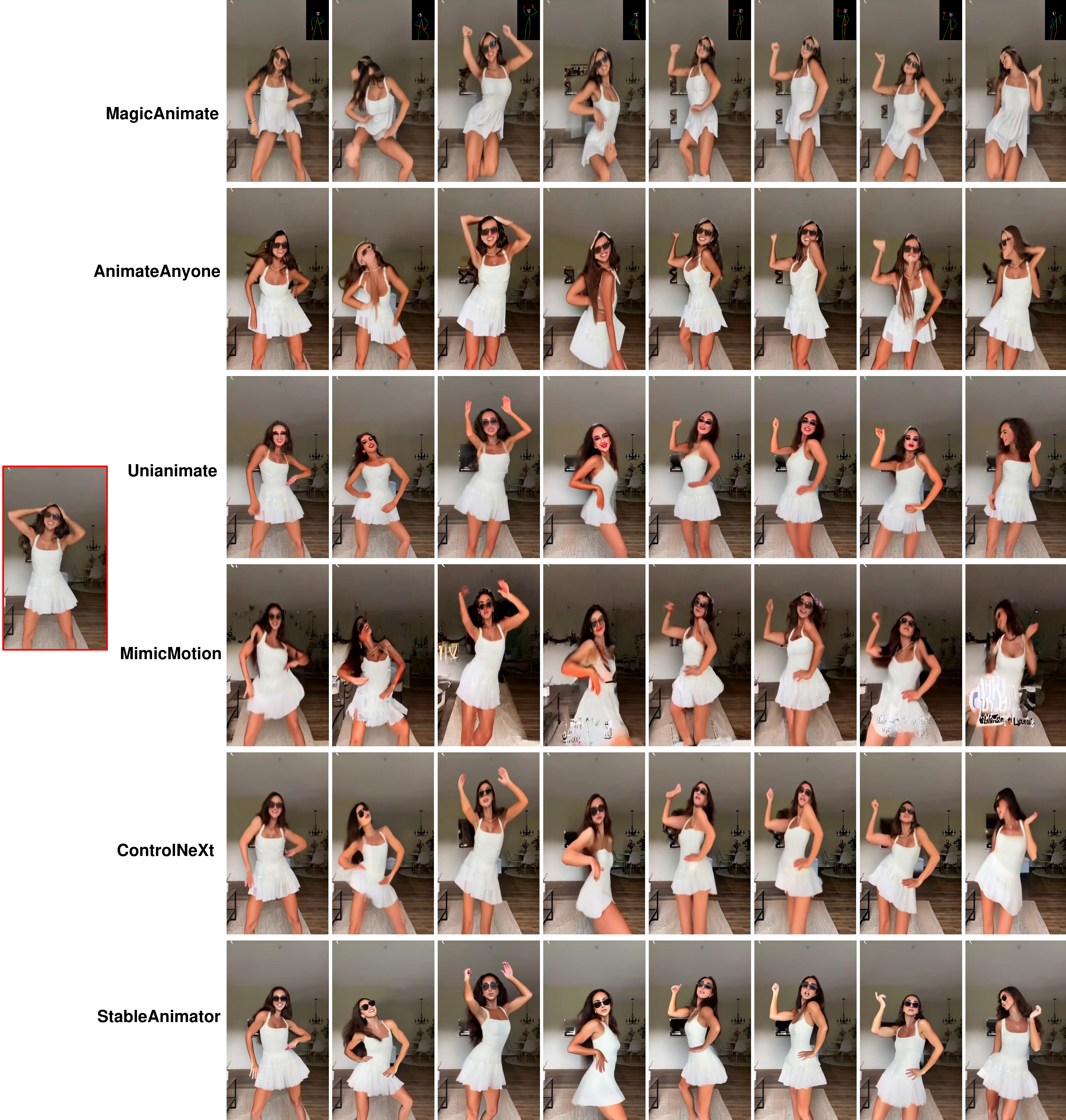}
\end{center}
\vspace{-0.55cm}
   \caption{Long animation results (3/3). The images with red borders are the reference images.}
\label{fig:long_video_3}
\vspace{-0.5cm}
\end{figure*}

We conduct several comparison experiments of our StableAnimator and SOTA human image animation models, as shown in Fig. \ref{fig:long_video_1}, Fig. \ref{fig:long_video_2}, and Fig. \ref{fig:long_video_3}. Each video contains more than 300 frames, featuring complex appearances of the protagonists, complicated motion sequences, and intricate background information. The results highlight the superiority of our StableAnimator in generating long animations while competing methods experience dramatic distortion of human bodies and identities.

\subsection{Multiple Person Animation}
\label{multiple_person}
\begin{figure*}[t!]
\begin{center}
\includegraphics[width=1\linewidth]{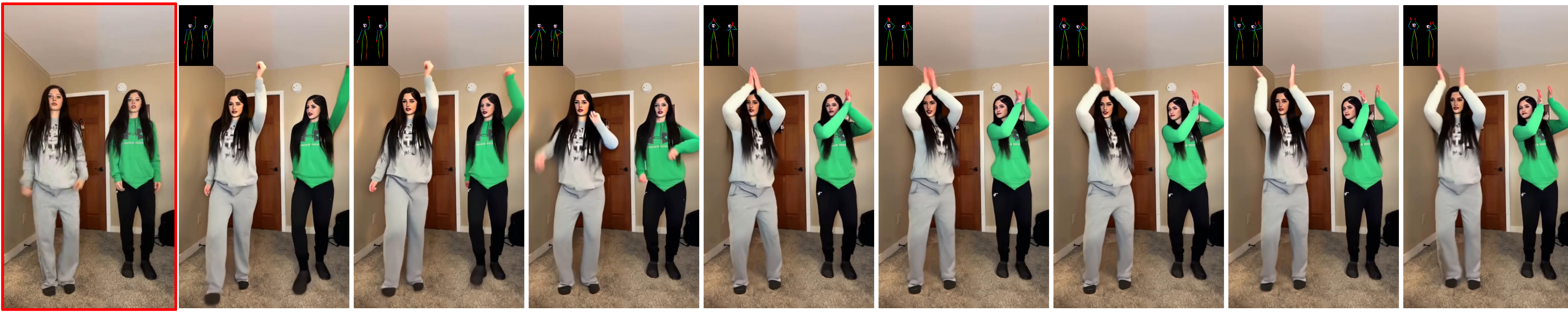}
\end{center}
\vspace{-0.45cm}
   \caption{Multiple-person animation results.}
\label{fig:multiple_person}
\end{figure*}

To demonstrate the robustness of our StableAnimator, we experiment on a particular video involving multiple protagonists, as shown in Fig. \ref{fig:multiple_person}. We can see that our StableAnimator is also capable of handling multiple-person animations while preserving the original identity and achieving high video fidelity.

\subsection{Optimization Details}
We present a more detailed HJB Equation-based Face Optimization in Algorithm \ref{alg:face_optimization}. Notably, the basic structure of our algorithm closely resembles Algorithm 2 in the EDM paper.
In the main paper, $\bm{\gamma}_{1}=-\bm{r}\cdot(\bm{X}_{1}-\bm{x}_{1})$ is derived from Eq.4 and Eq.5. In particular, this formula is obtained by calculating the transversality condition of Eq. 4 at the terminal time.

\subsection{Additional Face Discussion}
\begin{figure*}[t!]
\begin{center}
\includegraphics[width=1\linewidth]{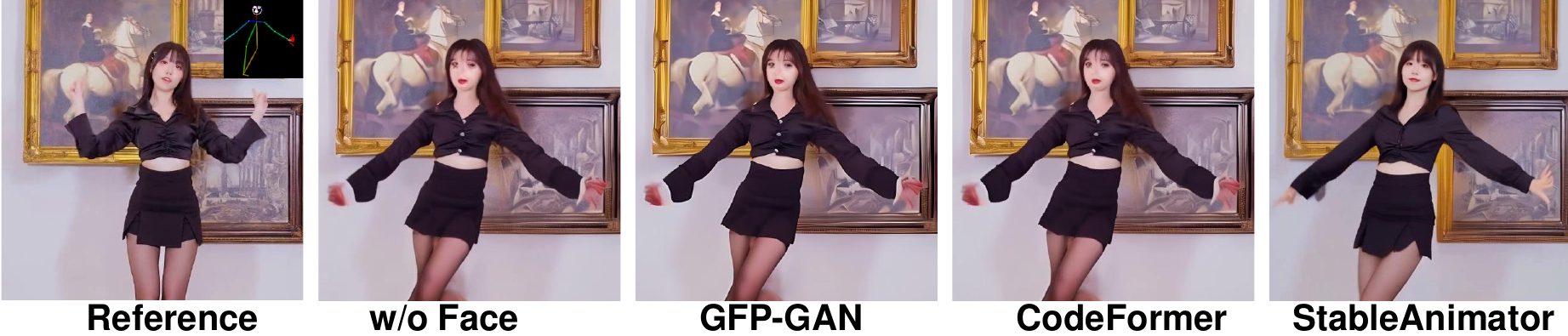}
\end{center}
\vspace{-0.45cm}
   \caption{Additional comparison results between our StableAnimator and current facial restoration models.}
\label{fig:face}
\vspace{-0.4cm}
\end{figure*}

We further conduct a comparison between our StableAnimator and other facial restoration models (GFP-GAN and CodeFormer). The results are shown in Fig. \ref{fig:face}. \textit{w/o} Face refers to the baseline model of our StableAnimator without incorporating any face-related components. It is noticeable that our StableAnimator has the best identity-preserving capability compared with other competitors, demonstrating the superiority of our StableAnimator regarding identity consistency. By contrast, GFP-GAN and CodeFormer suffer from serious facial distortion and over-sharpening. The plausible reason is that \textit{w/o} Face cannot synthesize the precise facial layout, which in turn undermines the effectiveness of subsequent facial restoration processes. This represents a fundamental limitation of post-processing-based face enhancement strategies.

\subsection{Identity-Preserving Loss}
In the image-domain identity-preserving methods, they often incorporate the ArcFace ID loss into the training process, which calculates the cosine similarity between the ArcFace face embeddings of the denoised result and the groundtruth. By contrast, during training, we introduce face masks extracted by Arcface to the conventional reconstruction MSE loss to improve modeling of face-related regions. The reason is that applying the ArcFace ID loss requires employing a VAE Decoder to convert the denoised latents into pixel level.
The reason is that applying the ArcFace ID loss requires using a VAE Decoder to convert the denoised latents into the pixel level. Although the VAE Decoder is frozen during training, a gradient back propagation graph must be maintained within the VAE Decoder to allow gradients to flow back to the U-Net for weight updates. However, the VAE Decoder in SVD contains memory-intensive temporal layers, making this back propagation graph extremely resource-demanding. 
Since training the SVD U-Net already requires substantial computational resources, incorporating the ArcFace ID loss would result in an unaffordable computational cost and significantly slow down the training process.
Therefore, we simply modify the reconstruction MSE loss by incorporating face masks to enable more explicit face modeling, making the training relatively lightweight.

\subsection{Additional Comparison Results}
\begin{figure*}[t!]
\begin{center}
\includegraphics[width=1\linewidth]{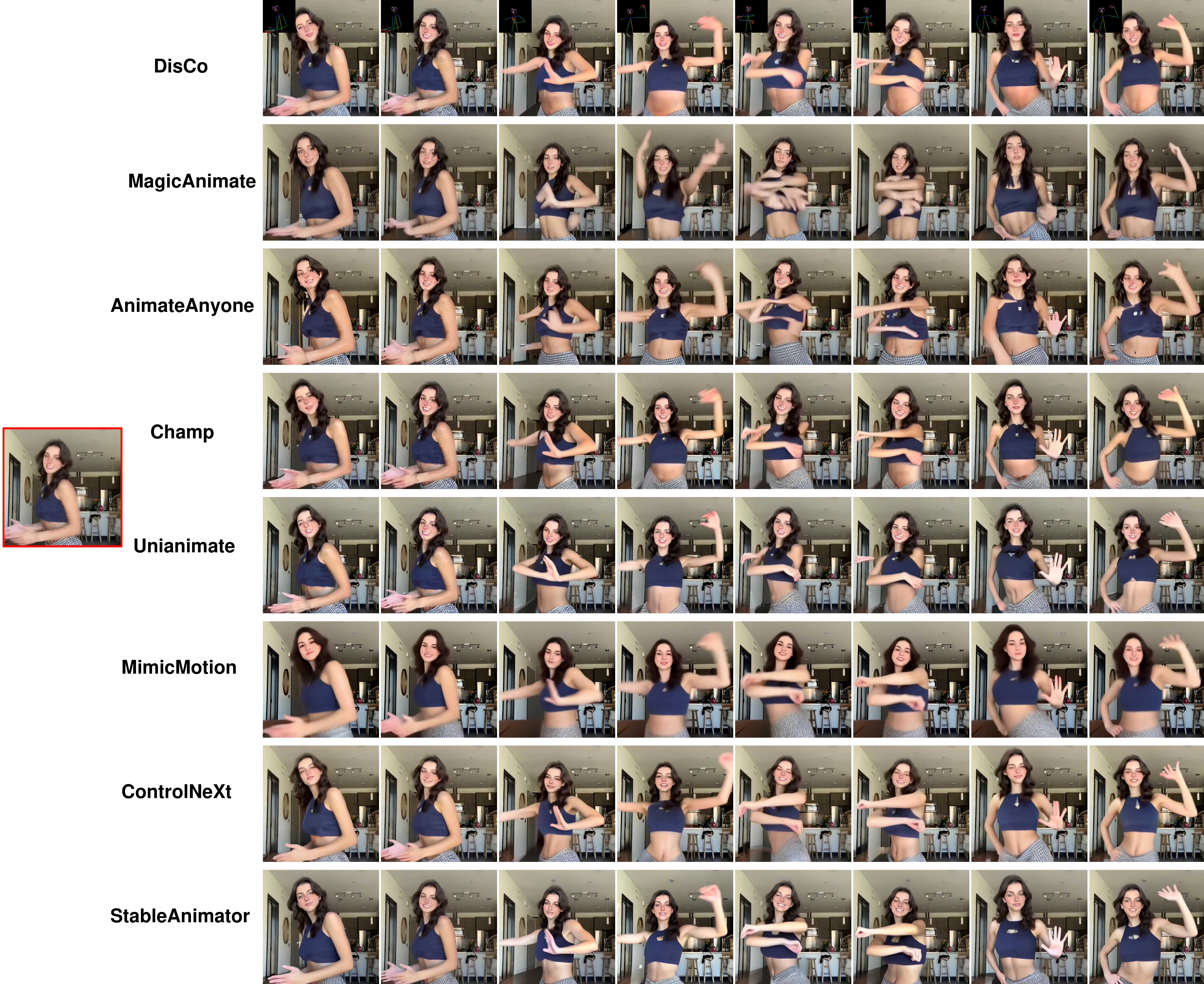}
\end{center}
\vspace{-0.55cm}
   \caption{Additional comparison results (1/2), using the case presented in the paper of MagicAnimate. The images with red borders are the reference images.}
\label{fig:more_comparison_1}
\vspace{-0.5cm}
\end{figure*}

\begin{figure*}[t!]
\begin{center}
\includegraphics[width=1\linewidth]{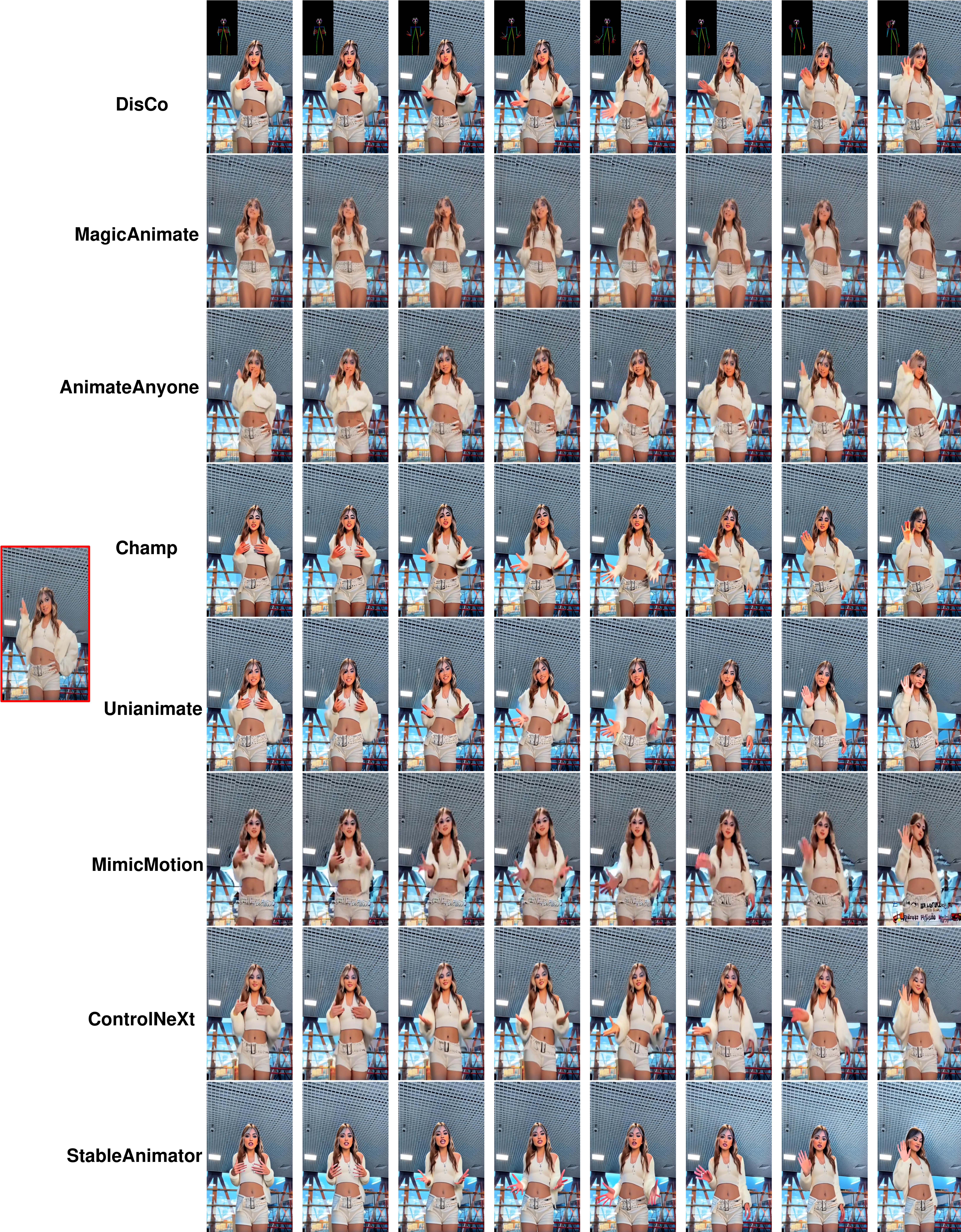}
\end{center}
\vspace{-0.55cm}
   \caption{Additional comparison results (2/2). The images with red borders are the reference images.}
\label{fig:more_comparison_2}
\vspace{-0.5cm}
\end{figure*}

Fig. \ref{fig:more_comparison_1} and Fig. \ref{fig:more_comparison_2} show additional comparison results. The provided pose sequences encompass complex motion information, and the initial poses of the reference images are two categories: one with the protagonist facing directly toward the camera, and another with the protagonist’s profile turned toward the camera. We can observe that our StableAnimator can accurately modify the motion of the reference images and maintain the original identity, while other competitors encounter varying degrees of human body distortion and loss of facial details.

\subsection{Animation Results}
\begin{figure*}[t!]
\begin{center}
\includegraphics[width=0.98\linewidth]{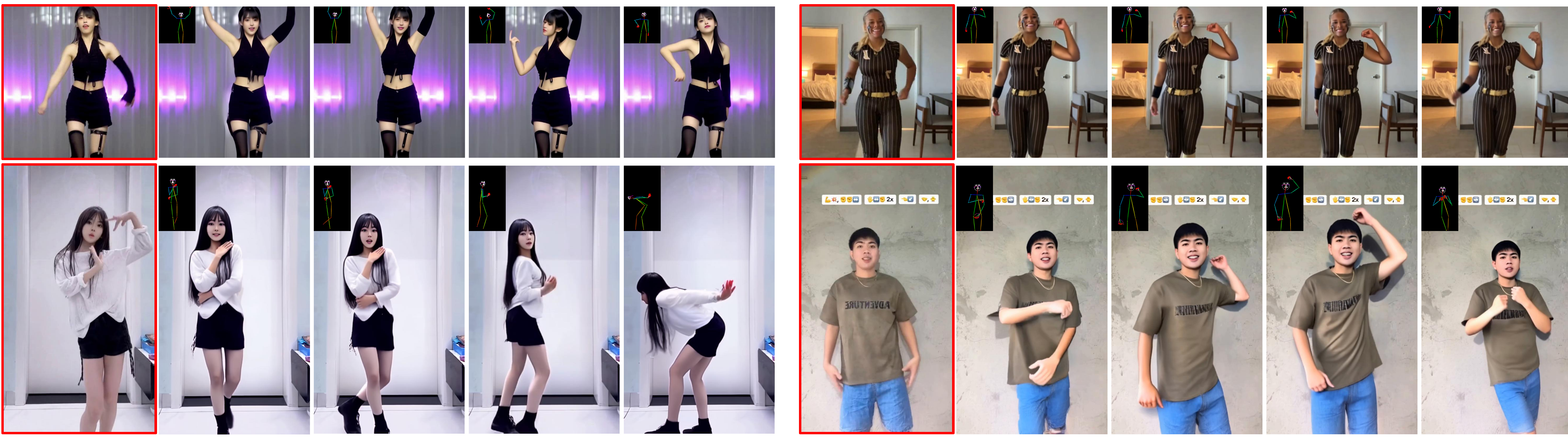}
\end{center}
\vspace{-0.55cm}
   \caption{Animation results of our StableAnimator.}
\label{fig:animation_results}
\vspace{-0.5cm}
\end{figure*}

\begin{figure*}[t!]
\begin{center}
\includegraphics[width=1\linewidth]{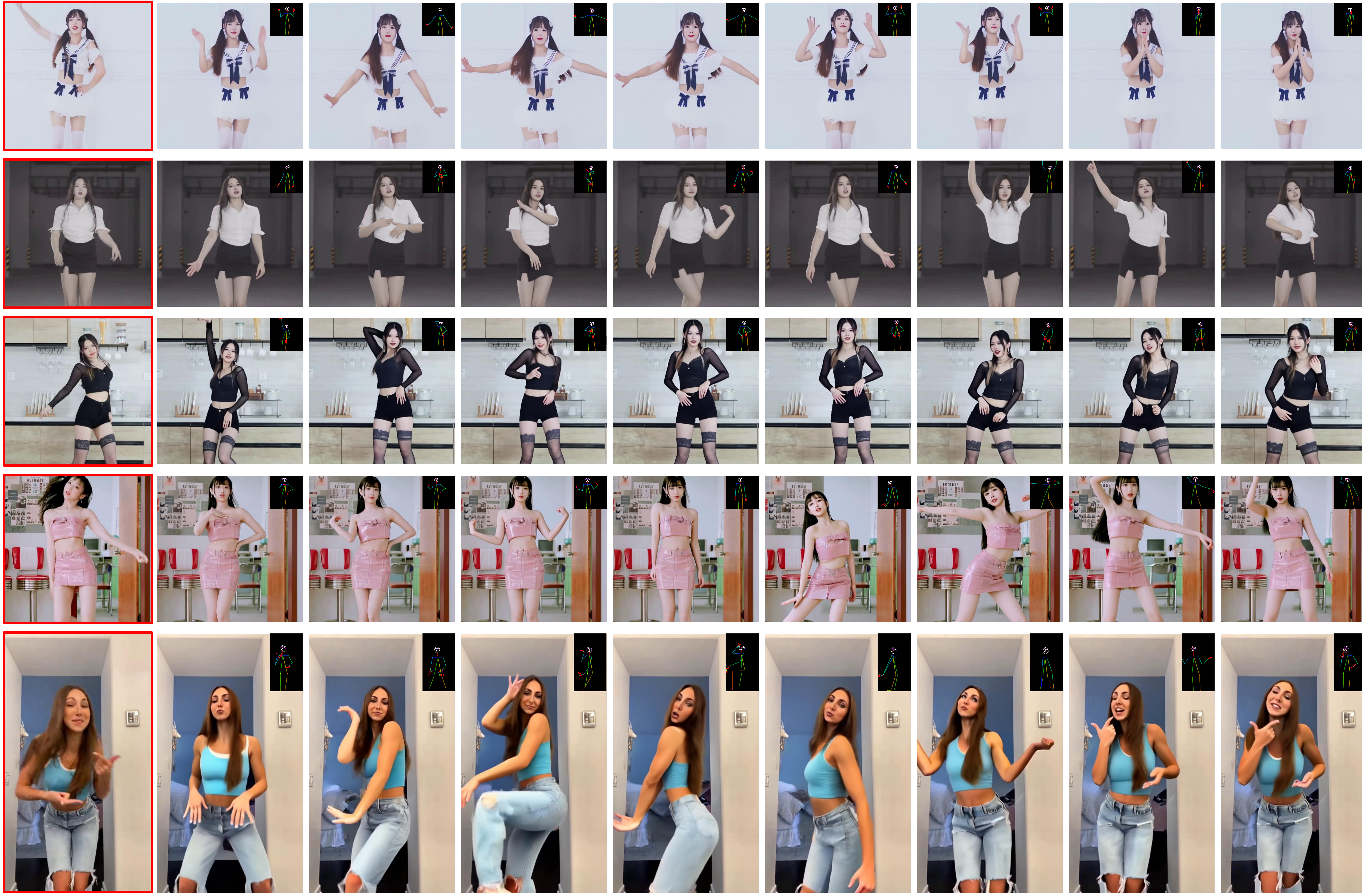}
\end{center}
\vspace{-0.55cm}
   \caption{Additional animation results (1/3). The images with red borders are the reference images.}
\label{fig:more_animations_1}
\vspace{-0.5cm}
\end{figure*}

\begin{figure*}[t!]
\begin{center}
\includegraphics[width=1\linewidth]{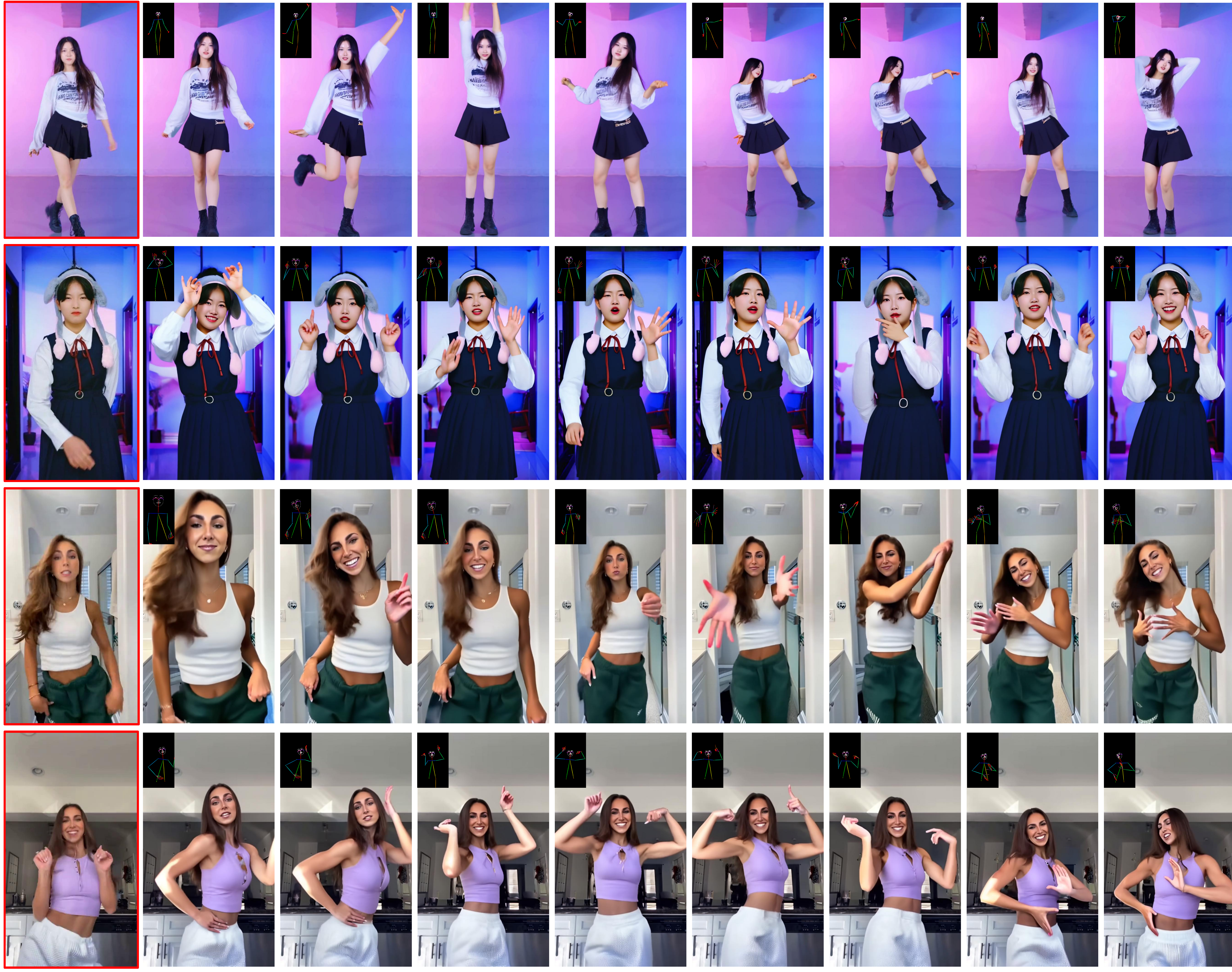}
\end{center}
\vspace{-0.55cm}
   \caption{Additional animation results (2/3). The images with red borders are the reference images.}
\label{fig:more_animations_2}
\vspace{-0.5cm}
\end{figure*}

\begin{figure*}[t!]
\begin{center}
\includegraphics[width=1\linewidth]{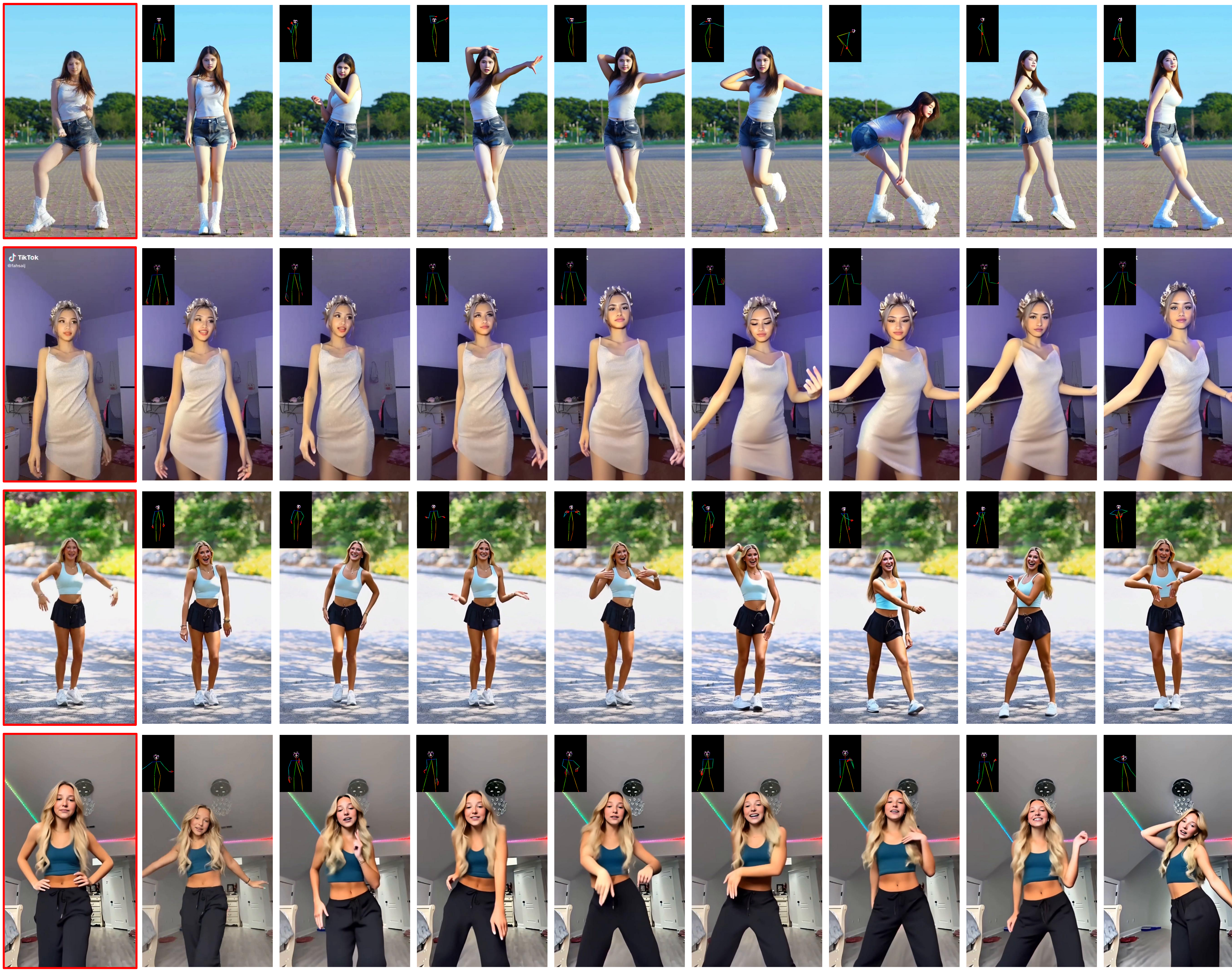}
\end{center}
\vspace{-0.55cm}
   \caption{Additional animation results (3/3). The images with red borders are the reference images.}
\label{fig:more_animations_3}
\vspace{-0.5cm}
\end{figure*}

We show our animation results in Fig. \ref{fig:animation_results}.
We can see that our StableAnimator can perform a wide range of human image animation while simultaneously preserving the protagonist’s appearance, background, and identity. 
Fig. \ref{fig:more_animations_1}, Fig. \ref{fig:more_animations_2}, and Fig. \ref{fig:more_animations_3} show additional animation results generated by our StableAnimator. 
Each cases contain complex protagonist's appearance and intricate motion information. For example, in the reference image in the fifth row of Fig. \ref{fig:more_animations_1}, the protagonist's closed eyes make it particularly challenging for the human animation model to preserve ID consistency.
It is noticeable that our StableAnimator can accurately manipulate motion in the reference image while preserving high-quality identity consistency, even in specific cases involving significant motion variations, such as head shaking and body rotation. Even when the head of the protagonist is continuously shaking and the angle facing the camera is constantly changing during the animation process, StableAnimator can still maintain a high level of identity consistency in the animation results without sacrificing details of the protagonist and the background.

\subsection{Additional Ablation Study}
\begin{figure*}[t!]
\begin{center}
\includegraphics[width=1\linewidth]{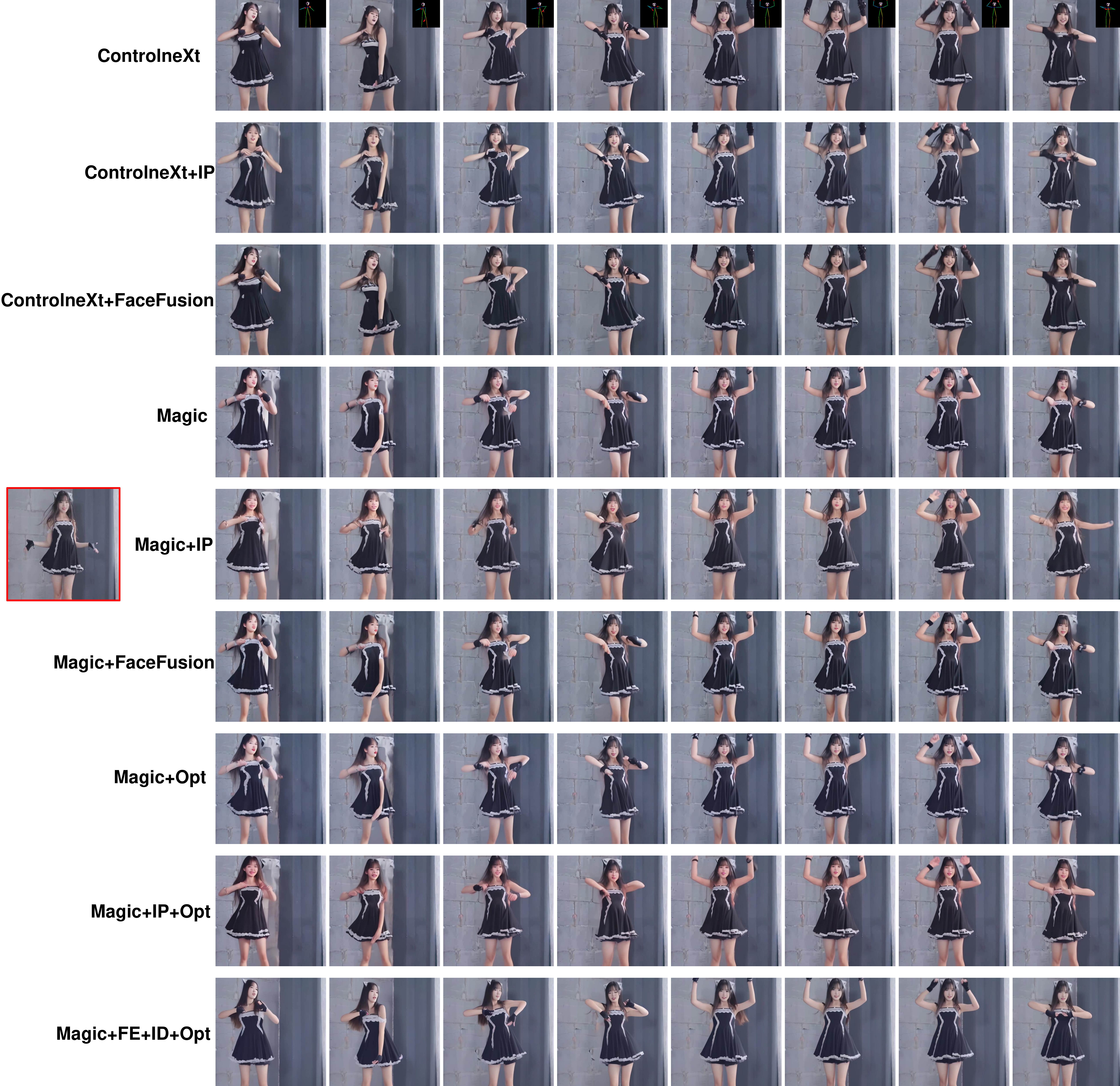}
\end{center}
\vspace{-0.55cm}
   \caption{Additional ablation study results. 
   IP, Magic, Opt, FE, and ID refer to IP-Adapter, MagicAnimate, our HJB Equation-based Face Optimization, our Global Content-Aware Face Encoder, and Distribution-Aware ID Adapter, respectively. 
   The images with red borders are the reference images.}
\label{fig:more_ablation}
\vspace{-0.5cm}
\end{figure*}

To validate the contribution of our proposed components, We conduct a more comprehensive qualitative ablation study on different diffusion backbones, as shown in Fig. \ref{fig:more_ablation}. ControlNeXt and MagicAnimate are based on Stable Video Diffusion (SVD) and Stable Diffusion (SD), respectively. We can see that our proposed components can significantly facilitate the performance of different backbone-based models, particularly in the facial regions. Notably, our proposed HJB Equation-based Face Optimization can still enhance the overall quality of animations to some extents, even when the backbone models lack any face-related encoders or adapters. The plausible reason is that our proposed HJB Equation-based Face Optimization can update the diffusion latents based on the face embedding similarity at each denoising step, thereby progressively refining the overall quality of denoised results without introducing any explicit face-related components.

\subsection{Limitation and Future Work}
\begin{figure*}[t!]
\begin{center}
\includegraphics[width=1\linewidth]{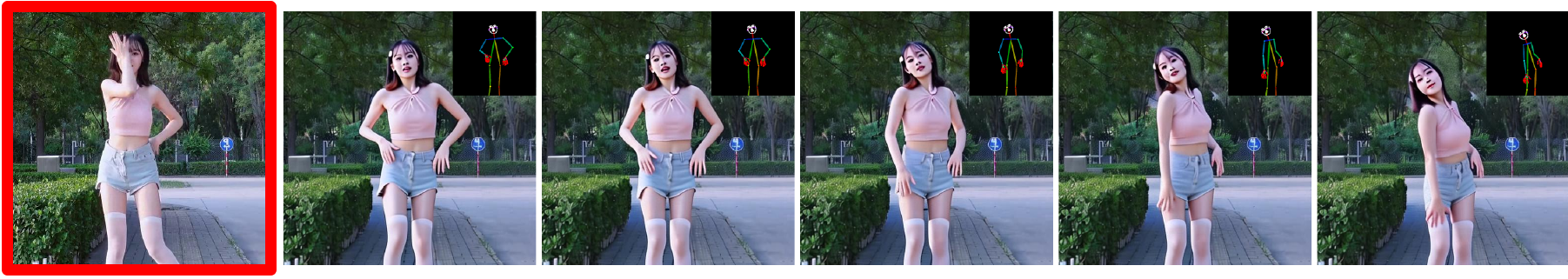}
\end{center}
\vspace{-0.55cm}
   \caption{One failure case of our StableAnimator.}
\label{fig:limitation}
\vspace{-0.5cm}
\end{figure*}

Fig. \ref{fig:limitation} shows one failure case of our StableAnimator. In the given reference image, the girl’s hand covers most of her face. Our StableAnimator struggles to fill in the obscured face regions, thereby degrading the quality of the synthesized face. One potential solution is introducing an additional face-aware inpainting adapter to the diffusion backbone for refining the face quality of given reference images. This part is left as future work. 

\subsection{Ethical Concern}
Our StableAnimator can animate the given reference image based on the given pose sequence, which can be implemented in various fields, including virtual reality and digital human creation. However, the potential misuse of this model, particularly for creating misleading content on social media platforms, is a concern. To mitigate this, it is essential to use sensitive content detection algorithms.

\end{document}